\documentclass[letterpaper]{article} 
\usepackage{aaai23}  
\usepackage{times}  
\usepackage{helvet}  
\usepackage{courier}  
\usepackage[hyphens]{url}  
\usepackage{graphicx} 
\usepackage{natbib}  
\usepackage{caption} 
\usepackage{xcolor}    
\definecolor{nicecolor1}{RGB}{70,130, 180}
\definecolor{nicecolor2}{RGB}{0, 111, 136}
\usepackage{algorithm}
\usepackage{algorithmic}
\usepackage[colorlinks, linkcolor=red, anchorcolor=green, citecolor=nicecolor2]{hyperref}  
\usepackage{amsmath}
\usepackage{multirow}
\usepackage{multicol}
\usepackage{threeparttable}
\usepackage{bbding}

\usepackage{newfloat}
\usepackage{listings}
\DeclareCaptionStyle{ruled}{labelfont=normalfont,labelsep=colon,strut=off} 
\floatstyle{ruled}
\newfloat{listing}{tb}{lst}{}
\floatname{listing}{Listing}
\usepackage{amsmath}
\title{Supplementary Materials for \\
``SKDBERT: Compressing BERT via Stochastic Knowledge Distillation"}
\author{
    Zixiang~Ding, \textsuperscript{\thanks{Corresponding author}{\rm 1}}
    Guoqing~Jiang, \textsuperscript{\rm 1}
    Shuai~Zhang, \textsuperscript{\rm 1}
    Lin~Guo, \textsuperscript{\rm 1}
    Wei~Lin \textsuperscript{\rm 2}
}
\affiliations{
    \textsuperscript{\rm 1}Meituan\\
    \textsuperscript{\rm 2}Individual\\

   \{dingzixiang, jiangguoqing03, zhangshuai51, guolin08\}@meituan.com, lwsaviola@163.com 
}

\usepackage{bibentry}

\begin{document}

\maketitle

\section{Multi-teacher Knowledge Distillation}
\label{sec:multi_teacher_kd}

For BERT-style language model compression, we verify the performance of MTBERT \citep{wu2021one} whose objection function can be expressed as:
\begin{align}
\mathcal{L}_{\rm MTBERT} = \sum_{i=1}^{N}\frac{{\rm CE}(y_i/T, y_s/T)}{1+{\rm CE}(y, y_i)},
\label{eq:l2}
\end{align}
where, $N$ indicates the number of teacher models, ${\rm CE}(\cdot, \cdot)$ is the cross-entropy loss, $T$ denotes the temperature, $y$ represents the ground-truth label, $y_i$ and $y_s$ refer to the outputs of $i$-th teacher model and the student model, respectively. We employ the student model used in TinyBERT \citep{jiao2020tinybert} which can be downloaded from \href{https://huggingface.co/huawei-noah/TinyBERT\_General\_6L\_768D}{\texttt{https://huggingface.co/huawei-noah/TinyB ERT\_General\_6L\_768D}}. Furthermore, we choose five BERT-style models whose architecture information is shown in Table \ref{tab:mtkd_teacher}, as the teacher models from the BERT official website\footnote{\href{https://github.com/google-research/bert}{\texttt{https://github.com/google-research/bert}}}.

\begin{table}[!h]
\centering
\small
\begin{threeparttable}[t]
\captionsetup{font={small}} 
\caption{The architecture information of each teacher model.}
\label{tab:mtkd_teacher}
\begin{tabular}{llcccc}
\hline
\textbf{Teacher} & \textbf{Layer} & \textbf{Hidden size} & \textbf{Head} & \textbf{\#Params (M)} \\
\hline
T$_{0}$ & 8   & 768 & 12 & 81.1 \\
T$_{1}$ & 10 & 768 & 12 & 95.3 \\
T$_{2}$ & 12 & 768 & 12 & 110 \\
T$_{3}$ & 24 & 1024 & 16 & 335 \\
T$_{4}$$$\dag$$ & 24 & 1024 & 16 & 335 \\
\hline
\end{tabular}
\scriptsize
\begin{tablenotes}
\item[$\dag$] Pre-training with whole word masking.
\end{tablenotes}
\end{threeparttable}
\end{table}   

Furthermore, we fine tune the above teacher models on six downstream tasks of GLUE benchmark, i.e., MRPC, RTE, SST-2, QQP, QNLI and MNLI whose detailed introduction can be found in Section \ref{sec:dataset_detail}. Moreover, the experimental settings of teacher models fine-tuning are shown as follows:
\begin{itemize}
\item \textbf{Learning Rate}: For T$_3$ and T$_4$, \{6e-6, 7e-6, 8e-6, 9e-6\} on MRPC and RTE tasks, and \{2e-5, 3e-5, 4e-5, 5e-5\} on other tasks. For other teacher models, \{2e-5, 3e-5, 4e-5, 5e-5\} on all tasks.
\item \textbf{Batch Size}: \{32\} for teacher model fine-tuning on each task.
\item \textbf{Epoch}: 10 for MRPC and RTE, 3 for other tasks.
\end{itemize}
Other settings are following TinyBERT \citep{jiao2020tinybert}.

Subsequently, we employ the ensemble of the fine-tuned teacher models to compress the student model via (\ref{eq:l2}) . Particularly, we only use the weighted multi-teacher distillation loss without the multi-teacher hidden loss and the task-specific loss as in MT-BERT \citep{wu2021one}.

The experimental settings are given as follows:
\begin{itemize}
\item \textbf{Learning Rate}: \{1e-5, 2e-5, 3e-5\} for all tasks.
\item \textbf{Batch Size}: \{16, 32, 64\}.
\item \textbf{Epoch}: 10 for MRPC and RTE, 3 for other tasks.
\end{itemize}
Similarly, other settings are also following TinyBERT \citep{jiao2020tinybert}.

\section{SKD for Image Classification}
\label{sec:skd_image}

To verify the effectiveness of \textit{general} SKD on computer vision, we conduct a list of experiments on CIFAR-100  \citep{krizhevsky2009learning} image classification dataset. Moreover, we employ ResNet \cite{he2016deep} and Wide ResNet (WRN) \cite{zagoruyko2016wide} with various architectures as the candidates of teacher team to distill a student model.  Following the experimental settings in CRD \cite{tian2020contrastive}, we train the student model and teacher model for 240 epochs, and employ SGD as the optimizer with a batch size of 64, a learning rate of 0.05 which is decayed by a factor of 0.1 when arriving at 150-th, 180-th, 210-th epoch and a weight decay of 5e-4, and show the results in Table \ref{tab:cv_acc_teacher_distillation}. 

Similar to \citet{tian2020contrastive}, we show the test accuracy of the last epoch in Table \ref{tab:cifar100} for a fair comparison. The proposed SKD also achieves novel performance on CIFAR-100. Compared to state-of-the-art CRD, our approach achieves the best performance in five out of six groups of distillation experiments.

\section{Why Does SKD Work?}
\label{sec:convergence}

In this section, taking uniform distribution based SKD as an example, we discuss why does it work.

\begin{table}[!t]
\centering
\scriptsize
\setlength\tabcolsep{1pt}
\begin{threeparttable}[tbq]
\captionsetup{font={small}} 
\caption{Accuracy (\%) of each student model and each teacher model for the last epoch of vanilla training and average of the best epoch over 5 runs of distillation on the test set of CIFAR-100.}
\label{tab:cv_acc_teacher_distillation}
\begin{tabular}{l|c|ccccc}
\hline
\multirow{2}{*}{\textbf{Teacher}} & \multicolumn{6}{c}{\textbf{Student}} \\
\cline{2-7}
 & None & WRN-40-1 & WRN-16-2 & ResNet-20 & ResNet-32  & ResNet-8$\times$4 \\
\hline
None & - & 71.98 & 73.95 & 69.06 & 71.14 & 72.50 \\
\hline
WRN-16-2                & 73.95 & 73.85 &  \\
WRN-22-2                & 74.26 & 74.36 & 75.24 \\
WRN-28-2                & 74.94 & 74.28 & 75.25 \\
WRN-34-2                & 76.22 & 74.23 & 75.16 \\
WRN-40-2                & 75.61 & 73.57 & 75.23 \\
\hline
ResNet-26                 & 70.75 & &  & 69.96 &  & \\
ResNet-32                 & 71.10 & &  & 70.43 &  & \\
ResNet-38                 & 71.97 & &  & 70.84 & 72.83 & \\
ResNet-44                 & 72.14 & &  & 70.99 & 73.20 &  \\
ResNet-50                 & 72.49 & &  & 70.96 & 72.97 & \\
ResNet-56                 & 72.41 & &  & 71.08 & 73.21 &  \\
ResNet-68                 & 73.62 & &  & 71.03 & 73.82 &  \\
ResNet-80                 & 73.67 & &  & 70.94 & 73.29 &  \\
ResNet-92                 & 73.94 & &  & 71.06 & 73.44 &  \\
ResNet-104               & 73.54 & &  & 71.11 & 73.55 &  \\
ResNet-110               & 74.31 & &  & 71.01 & 73.57 &  \\
\hline
ResNet-14$\times$4 & 77.07 & &  &           &           & 75.20 \\
ResNet-20$\times$4 & 77.67 & &  &           &           & 74.43 \\
ResNet-26$\times$4 & 79.29 & &  &           &           & 73.95 \\
ResNet-32$\times$4 & 79.42 & &  &           &           & 73.60 \\
\hline
\end{tabular}
\end{threeparttable}
\end{table}

\begin{table}[!t]
\centering
\scriptsize
\setlength\tabcolsep{1pt}
\begin{threeparttable}[tbq]
\captionsetup{font={small}} 
\caption{Test accuracy (\%) of the proposed SKD and other popular distillation approaches on CIFAR-100. All experimental results are cited from \citet{tian2020contrastive}. Average of the last epoch over 5 runs.}
\label{tab:cifar100}
\begin{tabular}{lcccccc}
\hline
Student& WRN-16-2 & WRN-40-1 & ResNet-20 & ResNet-20   & ResNet-32   & ResNet-8$\times$4 \\
\hline
Teacher & WRN-40-2 & WRN-40-2 & ResNet-56 & ResNet-110 & ResNet-110 & ResNet-32$\times$4 \\
\hline
Acc-S & 73.26 & 71.98 & 69.06 & 69.06 & 71.14 & 72.50 \\
Acc-T & 75.61 & 75.61 & 72.34 & 74.31 & 74.31 & 79.42 \\
\hline
KD     & 74.92 & 73.54 & 70.66 & 70.67 & 73.08 & 73.33 \\
FitNet    & 73.58 & 72.24 & 69.21 & 68.99 & 71.06 & 73.50 \\
AT   & 74.08 & 72.77 & 70.55 & 70.22 & 72.31 & 73.44 \\
SP         & 73.83 & 72.43 & 69.67 & 70.04 & 72.69 & 72.94 \\
CC       & 73.56 & 72.21 & 69.63 & 69.48 & 71.48 & 72.97 \\
VID      & 74.11 & 73.30 & 70.38 & 70.16 & 72.61 & 73.09 \\
RKD    & 73.35 & 72.22 & 69.61 & 69.25 & 71.82 & 71.90 \\
PKT   & 74.54 & 73.45 & 70.34 & 70.25 & 72.61 & 73.64 \\
AB     & 72.50 & 72.38 & 69.47 & 69.53 & 70.98 & 73.17 \\
FT    & 73.25 & 71.59 & 69.84 & 70.22 & 72.37 & 72.86 \\
FSP                & 72.91 &    -     & 69.95 & 70.11 & 71.89 & 72.62 \\
NST             & 73.68 & 72.24 & 69.60 & 69.53 & 71.96 & 73.30 \\
CRD    & 75.48 & 74.14 & \textbf{71.16} & 71.46 & 73.48 & \textbf{75.51} \\
\hline
\hline
TT$\dag$ & TT$_1$ & TT$_2$ & TT$_3$ & TT$_4$ & TT$_5$ & TT$_6$ \\
\hline
SKD  & \textbf{75.52} & \textbf{74.63} & \textbf{71.16} & \textbf{71.47} & \textbf{73.92} & 74.79 \\
\hline
\end{tabular}
\scriptsize
\begin{tablenotes}
\item[$\dag$]
Teacher team list: \\
TT$_1$: WRN-22-2, WRN-28-2, WRN-34-2, WRN-40-2 \\
TT$_2$: WRN-16-2, WRN-22-2, WRN-28-2, WRN-34-2, WRN-40-2 \\
TT$_3$: ResNet-26, ResNet-32, ResNet-38, ResNet-44, ResNet-50, ResNet-56 \\
TT$_4$: ResNet-32, ResNet-38, ResNet-44, ResNet-50, ResNet-56, ResNet-68, ResNet-80, ResNet-92, ResNet-104, ResNet-110 \\
TT$_5$: ResNet-44, ResNet-50, ResNet-56, ResNet-68, ResNet-80, ResNet-92, ResNet-104, ResNet-110 \\
TT$_6$: ResNet-14$\times$4, ResNet-20$\times$4, ResNet-32$\times$4 \\
\end{tablenotes}
\end{threeparttable}
\end{table} 

\subsection{Preliminaries: Knowledge Distillation}

On classification task with $C$ categories, vanilla KD employs the output of final layer, logits $\pmb{z}=[z_1, z_2, \dots, z_C]^{\rm T}$, to yield soft targets $\pmb{p}=[p_1,p_2,\dots,p_C]^{\rm T}$ for knowledge transfer as
\begin{equation}
p_{i} = \frac{{\rm exp}(z_i/T)}{\sum_{j=1}^{C}{\rm exp}(z_j/T)}, \quad i=1,2,\dots,C
\nonumber
\label{softtarget}
\end{equation}
where, $T$ represents the temperature factor which is used to control the importance of $p_i$.

The total loss $\mathcal{L}$ with respect to input $\pmb{x}$ and weights of student model $\pmb{W}$, consists of distilled loss $\mathcal{L}_d$ and student loss $\mathcal{L}_s$ as
\begin{equation}
\mathcal{L}(\pmb{x}, \pmb{W}) = \alpha \times \mathcal{L}_d(p(\pmb{z}_t, T), p(\pmb{z}_s, T))+ \beta \times \mathcal{L}_s(\pmb{y}, p(\pmb{z}_s, T)),
\nonumber
\label{totalloss}
\end{equation}
where,
\begin{equation}
\mathcal{L}_d(p(\pmb{z}_t, T), p(\pmb{z}_s, T)) = \sum_{i=1}^{C}-p(z_{ti}, T){\rm ln}(p(z_{si}, T))
\nonumber
\label{distilledloss}
\end{equation}
and
\begin{equation}
\mathcal{L}_s(\pmb{y}, p(\pmb{z}_s, T)) = \sum_{i=1}^{C}-y_i{\rm ln}(p(z_{si}, T))
\nonumber
\label{studentloss}
\end{equation}
treat soft targets $\pmb{p}(\pmb{z}_t, T)$ and ground truth $\pmb{y}$ as their labels, respectively. Moreover, $\pmb{z}_{t}$ and $\pmb{z}_{s}$ represent the logits of teacher and student models, respectively. Particularly, an ensemble of multiple teacher models is used to provide 

\subsection{The Convergence of SKD}

In order to optimize $\pmb{W}$, we should evaluate the cross-entropy gradient of total loss with regard to $z_{si}$, $\frac{\partial \mathcal{L}(\pmb{x}, \pmb{W})}{\partial z_{si}}$ as
\begin{equation}
\alpha \times \frac{\partial \mathcal{L}_d(p(\pmb{z}_t, T), p(\pmb{z}_s, T))}{\partial z_{si}} + \beta \times \frac{\partial \mathcal{L}_s(\pmb{y}, p(\pmb{z}_s, T))}{\partial z_{si}}.
\label{gradient}
\end{equation}
Here, the second term of the right side of \eqref{gradient} of SKD is identical with vanilla KD, so that we only discuss the first term.

The cross-entropy gradient of distilled loss with regard to $z_{si}$ can be computed as
\begin{equation}
\begin{split}
&\frac{\partial \mathcal{L}_d(p(\pmb{z}_t, T), p(\pmb{z}_s, T))}{\partial z_{si}} \\
&=\frac{\partial \sum_{k=1}^{C}-p(z_{tk}, T){\rm ln}(p(z_{sk}, T))}{\partial z_{si}}\\
&=\frac{\partial \sum_{k=1}^{C}-p(z_{tk}, T){\rm ln}(p(z_{sk}, T))}{\partial {\rm ln}(p(z_{sj}, T))} \times \frac{\partial {\rm ln}(p(z_{sj}, T))}{\partial z_{si}}\\
&=-\sum_{k=1}^{C}p(z_{tk}, T)\frac{1}{p(z_{sk}, T)} \times \frac{\partial {\rm ln}(p(z_{sj}, T))}{\partial z_{si}}.
\end{split}
\label{gradientld}
\end{equation}
For $\frac{\partial {\rm ln}(p(z_{sj}, T))}{\partial z_{si}}$, there are two situations (i.e., $i=j$ and $i \neq j$) as
\begin{equation}
\left\{
\begin{split}
&\frac{1}{T}{\rm ln}(p(z_{si}, T)) \times (1-{\rm ln}(p(z_{si}, T))), \quad if \, \, \, i=j\\
&-\frac{1}{T}{\rm ln}(p(z_{si}, T) \times {\rm ln}(p(z_{sj}, T)). \quad \quad \, \, \, others\\
\end{split}
\right.
\label{ij}
\end{equation}
Subsequently, we replace $\frac{\partial {\rm ln}(p(z_{sj}, T))}{\partial z_{si}}$ in \eqref{gradientld} with \eqref{ij}, and it can be rewritten as
\begin{equation}
\begin{split}
&\frac{\partial \mathcal{L}_d(p(\pmb{z}_t, T), p(\pmb{z}_s, T))}{\partial z_{si}}\\
&=-\sum_{k=1}^{C}\frac{p(z_{tk}, T)}{{\rm ln}(p(z_{sk}, T))} \times  \frac{\partial {\rm ln}(p(z_{sj}, T))}{\partial z_{si}}\\
&=-\frac{p(z_{ti}, T)}{{\rm ln}(p(z_{si}, T))} \times \frac{{\rm ln}(p(z_{si}, T)) \times (1-{\rm ln}(p(z_{si}, T)))}{T}\\
&+\sum_{j \neq i}\left(\frac{p(z_{tj}, T)}{{\rm ln}(p(z_{sj}, T))} \times \frac{{\rm ln}(p(z_{si}, T)) \times {\rm ln}(p(z_{sj}, T))}{T}\right)\\
&=-\frac{p(z_{ti}, T)-p(z_{ti}, T){\rm ln}(p(z_{si}, T))}{T} \\
&+\frac{\sum_{j \neq i}p(z_{tj}, T){\rm ln}(p(z_{si}, T))}{T}\\
&=\frac{\sum_{j=1}^Cp(z_{tj}, T){\rm ln}(p(z_{si}, T))-p(z_{ti}, T)}{T}\\
&=\frac{{\rm ln}(p(z_{si}, T))-p(z_{ti}, T)}{T}.
\nonumber
\label{gradientld2}
\end{split}
\end{equation}

On the one hand, in vanilla multi-teacher KD framework with an ensemble of $N$ teacher models, $z_{ti}$ is the average of ensemble as
\begin{equation}
z_{ti}=\frac{1}{N}\sum_{n=1}^Nz_{ti}^{(n)},
\nonumber
\label{zti}
\end{equation}
so that its soft target can be computed as
\begin{equation}
p_{\rm Avg}(z_{ti}, T)=\frac{{\rm exp}(\frac{\frac{1}{N}\sum_{n=1}^Nz_{ti}^{(n)}}{T})}{\sum_{j=1}^C {\rm exp}(\frac{\frac{1}{N}\sum_{n=1}^Nz_{tj}^{(n)}}{T})}.
\nonumber
\label{pzta}
\end{equation}
On the other hand, SKD randomly employs a single teacher from $N$ teacher models to obtain the soft target at $n$-th iteration as
\begin{equation}
p_{\rm SKD}^{(n)}(z_{ti}, T)=\frac{{\rm exp}(\frac{z_{ti}^{(n)}}{T})}{\sum_{j=1}^C {\rm exp}(\frac{z_{tj}^{(n)}}{T})},
\nonumber
\label{pzts}
\end{equation}
where, the sampled probability of $\pmb{z}_{t}^{(n)}$ subjects to uniform distribution ${\rm U}(1,N)$.

We assume that the number of iterations is $L$. In the total training process of multi-teacher KD, the amount of gradient update on $z_i$ dubbed $G_{\rm Avg}^{(i)}$ can be represented as
\begin{equation}
\begin{split}
&\sum_{l=1}^{L}\frac{{\rm ln}(p(z_{si}^{(l)}, T))-\frac{{\rm exp}(\frac{\frac{1}{N}\sum_{n=1}^Nz_{ti}^{(n)}}{T})}{\sum_{j=1}^C {\rm exp}(\frac{\frac{1}{N}\sum_{n=1}^Nz_{tj}^{(n)}}{T})}}{T}\\
&=\frac{1}{T}\sum_{l=1}^{L}{\rm ln}(p(z_{si}^{(l)}, T))-\frac{L}{T} \times \underbrace{\frac{{\rm exp}(\frac{\frac{1}{N}\sum_{n=1}^Nz_{ti}^{(n)}}{T})}{\sum_{j=1}^C {\rm exp}(\frac{\frac{1}{N}\sum_{n=1}^Nz_{tj}^{(n)}}{T})}}_{a_i}.
\end{split}
\label{amountavg}
\end{equation}
Beyond that, the amount of gradient update of SKD on $z_i$ dubbed $G_{\rm SKD}^{(i)}$ can be obtained by
\begin{equation}
\begin{split}
G_{\rm SKD}^{(i)}&=\sum_{l=1}^{L}\frac{{\rm ln}(p(z_{si}^{(l)}, T))-\frac{{\rm exp}(\frac{z_{ti}^{(n)}}{T})}{\sum_{j=1}^C {\rm exp}(\frac{z_{tj}^{(n)}}{T})}}{T}\\
&=\frac{1}{T}\sum_{l=1}^{L}{\rm ln}(p(z_{si}^{(l)}, T))- \frac{L}{T} \times \underbrace{\frac{{\rm exp}(\frac{z_{ti}^{(n)}}{T})}{\sum_{j=1}^C {\rm exp}(\frac{z_{tj}^{(n)}}{T})}}_{b_i}
\end{split}
\label{amountsto}
\end{equation}

Obviously, the main difference between $G_{\rm Avg}^{(i)}$ and $G_{\rm SKD}^{(i)}$ is the terms of $a_i$ in \eqref{amountavg} and $b_i$ in \eqref{amountsto}. We denote $\frac{\sum_{n=1}^Nz_{ti}^{(n)}}{N}$ as $\overline{z}_{ti}$, and rewrite term $a_i$ in \eqref{amountavg} as
\begin{equation}
\begin{split}
a_i=\frac{{\rm exp}(\frac{\overline{z}_{ti}}{T})}{\sum_{j=1}^C {\rm exp}(\frac{\overline{z}_{tj}}{T})}.
\end{split}
\label{a}
\end{equation}
According to \eqref{a} and the term $b_i$ in \eqref{amountsto}, we can find that their relationship is similar to the relationship between Batch Gradient Descent (BGD) and Stochastic Gradient Descent (SGD). Moreover, BGD employs total data to obtain the gradient in each iteration, but SGD uses single one. Based on stochastic approximation (e.g. ROBBINS-MONRO's theory \cite{robbins1951stochastic}) and convex optimization, the convergence of SGD has been proven \cite{turinici2021convergence}.

\section{Details of GLUE Benchmark}
\label{sec:dataset_detail}

GLUE consists of 9 NLP tasks: Microsoft Research Paraphrase Corpus (MRPC) \cite{dolan2005automatically}, Recognizing Textual Entailment (RTE) \cite{bentivogli2009fifth}, Corpus of Linguistic Acceptability (CoLA) \cite{warstadt2019neural}, Semantic Textual Similarity Benchmark (STS-B) \cite{cer2017semeval}, Stanford Sentiment Treebank (SST-2) \cite{socher2013recursive}, Quora Question Pairs (QQP) \cite{chen2018quora}, Question NLI (QNLI) \cite{rajpurkar2016squad}, Multi-Genre NLI (MNLI) \cite{williams2017broad}, and Winograd NLI (WNLI) \cite{levesque2012winograd}.

\paragraph{MRPC}

belongs to a sentence similarity task where system aims to identify the paraphrase/semantic equivalence relationship between two sentences.

\paragraph{RTE}

belongs to a natural language inference task where system aims to recognize the entailment relationship of given two text fragments.

\paragraph{CoLA}

belongs to a single-sentence task where system aims to predict the grammatical correctness of an English sentence.

\paragraph{STS-B}

belongs to a sentence similarity task where system aims to evaluate the similarity of two pieces of texts by a score from 1 to 5. 

\paragraph{SST-2}

belongs to a single-sentence task where system aims to predict the sentiment of movie reviews.

\paragraph{QQP}

belongs to a sentence similarity task where system aims to identify the semantical equivalence of two questions from the website Quora.  

\paragraph{QNLI}

belongs to a natural language inference task where system aims to recognize that for a given pair \textless\textit{question}, \textit{context}\textgreater, the answer to the \textit{question} whether contains in the \textit{context}.   

\paragraph{MNLI}

belongs to a natural language inference task where system aims to predict the possible relationships (i.e., entailment, contradiction and neutral) of \textit{hypothesis} with regard to \textit{premise} for a given pair \textless\textit{premise}, \textit{hypothesis}\textgreater,. 

\paragraph{WNLI}

belongs to a natural language inference task where system aims to determine the referent of a sentence's pronoun from a list of choices.

\section{Evaluation Metrics and Hyper-parameters}
\label{sec:metrics_hp}

\subsection{Evaluation Metrics}
\label{sec:metrics}

In order to choose the best model, we use appropriate metrics on GLUE-dev as shown in Table \ref{tab:metrics}.

\begin{table}[!h]
\centering
\scriptsize
\setlength\tabcolsep{2pt}
\begin{threeparttable}[tbq]
\captionsetup{font={small}} 
\caption{Evaluation metrics on GLUE-dev.}
\label{tab:metrics}
\begin{tabular}{lll}
\hline
\textbf{Task} & \textbf{Metric} & \textbf{Description} \\
\hline
\multirow{2}{*}{STS-B} & \multirow{2}{*}{$\frac{pear+spear}{2}$} & $pear$ and $spear$ are Pearson and Spearman \\ &&correlation coefficients, respectively. \\
MNLI & $acc$-$m$ & $acc$-$m$ means accuracy on matched section. \\
MRPC, QQP &  $\frac{acc+f1}{2}$ & $acc$ represents accuracy, $f1$ indicates F1 scores. \\
SST-2, QNLI, RTE & $acc$ & None \\
\hline
\end{tabular}
\end{threeparttable}
\end{table}

\subsection{Fine-tuning Hyper-parameters}
\label{sec:hp}

In this paper, we need to perform teacher model fine-tuning, student model fine-tuning and student model distillation. For the above three cases, the used hyper-parameters are shown in Table \ref{tab:hp}.

\begin{table}[!h]
\centering
\scriptsize
\setlength\tabcolsep{13pt}
\begin{threeparttable}[tbq]
\captionsetup{font={small}} 
\caption{Hyper-parameters for fine-tuning and distillation.}
\label{tab:hp}
\begin{tabular}{ll}
\hline
\textbf{Hyper-parameter} & \textbf{Value} \\
\hline
\multirow{6}{*}{Learning rate} & For large teacher models, [6e-6, 7e-6, 8e-6, \\ 
                                                     & 9e-6] on MRPC and RTE tasks, [2e-5, 3e-5, \\
                                                     & 4e-5, 5e-5] on other tasks. For other teacher \\
                                                     & models and student models, [2e-5, 3e-5, 4e-\\
                                                     & 5, 5e-5] and [1e-5, 2e-5, 3e-5] on all tasks, \\
                                                     & respectively. \\
Adam $\epsilon$ &  1e-6 \\
Adam $\beta_1$ &  0.9 \\
Adam $\beta_2$ &  0.999 \\
Learning rate decay  &  linear \\
Warmup fraction  &  0.1 \\
Attention dropout & 0.1 \\
Dropout & 0.1\\
Weight decay  &  1e-4 \\
Batch size  &  32 for fine-tuning, [16, 32] for distillation\\
\multirow{4}{*}{Fine-tuning epochs} & For KD, WKD and TAKD, 10 on MRPC,\\
                        									   & RTE and STS-B tasks, 3 on other tasks. For\\
                        									   & SKD, 15 on MRPC, RTE and STS-B tasks,\\
                        									   & 5 on other tasks \\ 
\hline
\end{tabular}
\end{threeparttable}
\end{table}

\begin{table}[!h]
\centering
\scriptsize
\begin{threeparttable}[t]
\captionsetup{font={small}} 
\caption{The architecture of each student and teacher model.}
\label{tab:student_teacher_info}
\begin{tabular}{llcccc}
\hline
\textbf{Model} & \textbf{Name} & \textbf{Layer} & \textbf{Hidden Size} & \textbf{Head} & \textbf{\#Params (M)} \\
\hline
\multirow{2}{*}{Student} & SKDBERT$_4$ & 4 & 312 & 12 & 14.5\\
& SKDBERT$_6$ & 6 & 768 & 12 & 67.0\\
\hline
\hline
\multirow{14}{*}{Teacher} 
& T$_{01}$   & 8   & 128 & 2 & 5.6 \\
& T$_{02}$   & 10 & 128 & 2 & 6.0 \\
& T$_{03}$   & 12 & 128 & 2 & 6.4 \\
& T$_{04}$   & 8   & 256 & 4 & 14.3 \\
& T$_{05}$   & 10 & 256 & 4 & 15.9 \\
& T$_{06}$   & 12 & 256 & 4 & 17.5 \\
& T$_{07}$   & 8   & 512 & 8 & 41.4 \\
& T$_{08}$   & 10 & 512 & 8 & 47.7 \\
& T$_{09}$   & 12 & 512 & 8 & 54.0 \\
& T$_{10}$ & 8   & 768 & 12 & 81.1 \\
& T$_{11}$ & 10 & 768 & 12 & 95.3 \\
& T$_{12}$ & 12 & 768 & 12 & 110 \\
& T$_{13}$ & 24 & 1024 & 16 & 335 \\
& T$_{14}$$$\dag$$ & 24 & 1024 & 16 & 335 \\
\hline
\end{tabular}
\scriptsize
\begin{tablenotes}
\item[$\dag$] Pre-training with whole word masking.
\end{tablenotes}
\end{threeparttable}
\end{table}

\section{Student Model and Teacher Team Details}
\label{sec:student_teacher_details}

The architecture information of each student model and teacher model is shown in Table \ref{tab:student_teacher_info}. On the one hand, we direct treating the pre-trained model of TinyBERT$_4$\footnote{\href{https://huggingface.co/huawei\-noah/TinyBERT\_General\_4L\_312D}{\texttt{https://huggingface.co/huawei-noah/TinyBE RT\_General\_4L\_312D}}} and TinyBERT$_6$\footnote{\href{https://huggingface.co/huawei-noah/TinyBERT\_General\_6L\_768D}{\texttt{https://huggingface.co/huawei-noah/TinyBE RT\_General\_6L\_768D}}} as the student models of SKDBERT. On the other hand, we choose 14 BERT models with various capabilities as the candidates for teacher team. Moreover, each pre-trained teacher model can be downloaded from official implementation of BERT\footnote{\href{https://github.com/google-research/bert}{\texttt{https://github.com/google-research/bert}}}. Furthermore, the results of the student model and the teacher model on GLUE-dev are shown in Table \ref{tab:student_teacher_ft} which can be used to obtain the teacher-rank sampling distribution.

\begin{table}[!t]
\centering
\scriptsize
\setlength\tabcolsep{3pt}
\begin{threeparttable}[tbq]
\captionsetup{font={small}} 
\caption{The fine-tuning performance of each student and teacher models on GLUE-dev.}
\label{tab:student_teacher_ft}
\begin{tabular}{lccccccccc}
\hline
\textbf{Model} & \textbf{MRPC} & \textbf{RTE} &  \textbf{STS-B} & \textbf{SST-2} & \textbf{QQP} & \textbf{QNLI} & \textbf{MNLI-m} & \textbf{Avg}\\
\hline
SKDBERT$_4$   & 83.83 & 63.90 & 84.86 & 87.84 & 84.64 & 83.83 & 76.18 & 80.73 \\
SKDBERT$_6$   & 89.44 & 71.84 & 88.52 & 91.63 & 86.44 & 90.70 & 82.55 & 85.87 \\
\hline
\hline
T$_{01}$   & 81.83 & 66.06 & 85.15 & 86.24 & 83.95 & 83.80 & 72.95 & 80.00 \\
T$_{02}$   & 84.75 & 66.06 & 84.99 & 85.67 & 84.18 & 84.00 & 73.75 & 80.49 \\
T$_{03}$   & 84.59 & 65.70 & 85.73 & 86.47 & 85.02 & 84.40 & 75.16 & 81.01 \\
T$_{04}$   & 85.18 & 64.62 & 86.77 & 89.33 & 86.36 & 86.80 & 78.16 & 82.46 \\
T$_{05}$   & 87.84 & 66.06 & 87.45 & 89.33 & 87.25 & 87.26 & 78.75 & 83.42 \\
T$_{06}$   & 85.96 & 66.06 & 87.00 & 89.68 & 87.21 & 87.42 & 79.54 & 83.27 \\
T$_{07}$   & 87.91 & 70.04 & 88.46 & 91.28 & 88.69 & 89.27 & 80.84 & 85.21 \\
T$_{08}$   & 88.17 & 65.70 & 88.75 & 91.28 & 88.62 & 89.25 & 81.41 & 84.74 \\
T$_{09}$   & 88.85 & 66.43 & 88.74 & 92.09 & 89.01 & 90.33 & 81.90 & 85.34 \\
T$_{10}$   & 89.36 & 68.95 & 89.03 & 93.00 & 89.27 & 90.79 & 83.05 & 86.21 \\
T$_{11}$   & 90.10 & 71.12 & 89.59 & 92.78 & 89.71 & 91.20 & 84.00 & 86.93 \\
T$_{12}$   & 89.98 & 68.59 & 90.20 & 92.66 & 89.66 & 91.85 & 84.40 & 86.76\\
T$_{13}$   & \textbf{90.60} & 62.74 & 90.13 & 94.50 & 90.26 & 92.70 & 86.88 & 86.83 \\
T$_{14}$   & 90.15 & \textbf{79.06 }& \textbf{91.21} & \textbf{94.72} & \textbf{90.40} & \textbf{93.89} & \textbf{87.06} & \textbf{89.50}\\
\hline
\end{tabular}
\end{threeparttable}
\end{table}

\section{Distillation Performance of SKDBERT with Various Teacher Models}
\label{sec:student_distillation_performance}

In order to determine the student-rank sampling distribution for SKD, we employ each teacher model to distill SKDBERT$_4$ and SKDBERT$_6$ on GLUE-dev, and show the results in Table \ref{tab:student_kd}. 

\begin{table}[!t]
\centering
\tiny
\setlength\tabcolsep{1pt}
\begin{threeparttable}[tbq]
\captionsetup{font={small}} 
\caption{The distillation performance of SKDBERT with various teacher models on GLUE-dev. \textcolor{red}{$\uparrow$} indicates that the distillation performance is better than SKDBERT without KD.  \textcolor{green}{$\downarrow$} indicates that the distillation performance is worse than SKDBERT without KD.}
\label{tab:student_kd}
\begin{tabular}{llccccccc}
\hline
\textbf{Student} & \textbf{Teacher} &  \textbf{MRPC} & \textbf{RTE} & \textbf{STS-B} & \textbf{SST-2} & \textbf{QQP} & \textbf{QNLI} & \textbf{MNLI-m} \\
\hline
SKDBERT$_4$ & None & 83.83 & 63.90 & 84.86 & 87.84 & 84.64 & 83.83 & 76.18 \\
\hline
\multirow{14}{*}{SKDBERT$_4$} 
& T$_{01}$   & 81.73 (\textcolor{green}{$\downarrow$}) & 62.82 (\textcolor{green}{$\downarrow$}) & 84.90 (\textcolor{red}{$\uparrow$}) & 86.93 (\textcolor{green}{$\downarrow$}) & 83.54 (\textcolor{green}{$\downarrow$})  & 83.60 (\textcolor{green}{$\downarrow$}) & 73.39 (\textcolor{green}{$\downarrow$}) \\
& T$_{02}$   & 83.55 (\textcolor{green}{$\downarrow$}) & 64.62 (\textcolor{red}{$\uparrow$}) & 84.63 (\textcolor{green}{$\downarrow$}) & 87.39 (\textcolor{green}{$\downarrow$}) & 84.01 (\textcolor{green}{$\downarrow$})  & 83.82 (\textcolor{green}{$\downarrow$}) & 73.81 (\textcolor{green}{$\downarrow$}) \\
& T$_{03}$   & 83.88 (\textcolor{red}{$\uparrow$}) & 65.34 (\textcolor{red}{$\uparrow$}) & 84.91 (\textcolor{red}{$\uparrow$}) & 87.73 (\textcolor{green}{$\downarrow$}) & 84.75 (\textcolor{red}{$\uparrow$}) & 83.75 (\textcolor{green}{$\downarrow$}) & 74.90 (\textcolor{green}{$\downarrow$}) \\
& T$_{04}$   & 82.73 (\textcolor{green}{$\downarrow$}) & 63.90 (\textcolor{red}{$\uparrow$}) & 84.74 (\textcolor{green}{$\downarrow$}) & 88.07 (\textcolor{red}{$\uparrow$}) & 85.26 (\textcolor{red}{$\uparrow$}) & 85.23 (\textcolor{red}{$\uparrow$}) & 76.58 (\textcolor{red}{$\uparrow$}) \\
& T$_{05}$   & 85.25 (\textcolor{red}{$\uparrow$}) & \textbf{67.15} (\textcolor{red}{$\uparrow$}) & 84.66 (\textcolor{green}{$\downarrow$}) & 87.84 (\textcolor{red}{$\uparrow$}) & 85.48 (\textcolor{red}{$\uparrow$}) & 85.34 (\textcolor{red}{$\uparrow$}) & 76.94 (\textcolor{red}{$\uparrow$}) \\
& T$_{06}$   & \textbf{86.25} (\textcolor{red}{$\uparrow$}) & 66.06 (\textcolor{red}{$\uparrow$}) & 85.05 (\textcolor{red}{$\uparrow$}) & 87.50 (\textcolor{green}{$\downarrow$}) & 85.40 (\textcolor{red}{$\uparrow$}) & \textbf{85.94} (\textcolor{red}{$\uparrow$}) & 77.30 (\textcolor{red}{$\uparrow$}) \\
& T$_{07}$   & 85.31 (\textcolor{red}{$\uparrow$}) & 64.98 (\textcolor{red}{$\uparrow$}) & 85.11 (\textcolor{red}{$\uparrow$}) & 88.53 (\textcolor{red}{$\uparrow$}) & 85.32 (\textcolor{red}{$\uparrow$}) & 84.77 (\textcolor{red}{$\uparrow$}) & 77.53 (\textcolor{red}{$\uparrow$}) \\
& T$_{08}$   & 84.27 (\textcolor{red}{$\uparrow$}) & 65.34 (\textcolor{red}{$\uparrow$}) & \textbf{85.17} (\textcolor{red}{$\uparrow$}) & 88.19 (\textcolor{red}{$\uparrow$}) & \textbf{85.59} (\textcolor{red}{$\uparrow$}) & 85.52 (\textcolor{red}{$\uparrow$}) & \textbf{77.84} (\textcolor{red}{$\uparrow$}) \\
& T$_{09}$   & 84.01 (\textcolor{red}{$\uparrow$}) & 66.79 (\textcolor{red}{$\uparrow$}) & 84.67 (\textcolor{green}{$\downarrow$}) & 88.30 (\textcolor{red}{$\uparrow$}) & 85.39 (\textcolor{red}{$\uparrow$}) & 85.63 (\textcolor{red}{$\uparrow$}) & 77.15 (\textcolor{red}{$\uparrow$}) \\
& T$_{10}$  & 84.87 (\textcolor{red}{$\uparrow$}) & 65.70 (\textcolor{red}{$\uparrow$}) & 84.55 (\textcolor{green}{$\downarrow$}) & 87.96 (\textcolor{red}{$\uparrow$}) & 85.30 (\textcolor{red}{$\uparrow$}) & 85.37 (\textcolor{red}{$\uparrow$}) & 77.82 (\textcolor{red}{$\uparrow$}) \\
& T$_{11}$  & 83.96 (\textcolor{red}{$\uparrow$}) & 65.34 (\textcolor{red}{$\uparrow$}) & 84.96 (\textcolor{red}{$\uparrow$}) & 88.65 (\textcolor{red}{$\uparrow$}) & 85.44 (\textcolor{red}{$\uparrow$}) & 85.04 (\textcolor{red}{$\uparrow$}) & 77.36 (\textcolor{red}{$\uparrow$}) \\
& T$_{12}$  & 85.24 (\textcolor{red}{$\uparrow$}) & 65.70 (\textcolor{red}{$\uparrow$}) & 85.03 (\textcolor{red}{$\uparrow$}) & \textbf{89.22} (\textcolor{red}{$\uparrow$}) & 85.32 (\textcolor{red}{$\uparrow$}) & 85.58 (\textcolor{red}{$\uparrow$}) & 77.40 (\textcolor{red}{$\uparrow$}) \\
& T$_{13}$  & 85.35 (\textcolor{red}{$\uparrow$}) & 66.43 (\textcolor{red}{$\uparrow$}) & 84.87 (\textcolor{red}{$\uparrow$}) & 88.30 (\textcolor{red}{$\uparrow$}) & 85.13 (\textcolor{red}{$\uparrow$}) & 84.88 (\textcolor{red}{$\uparrow$}) & 77.08 (\textcolor{red}{$\uparrow$}) \\
& T$_{14}$  & 84.42 (\textcolor{red}{$\uparrow$}) & 66.43 (\textcolor{red}{$\uparrow$}) & 84.87 (\textcolor{red}{$\uparrow$}) & 87.73 (\textcolor{green}{$\downarrow$}) & 85.29 (\textcolor{red}{$\uparrow$}) & 84.51 (\textcolor{red}{$\uparrow$}) & 76.73 (\textcolor{red}{$\uparrow$}) \\
\hline
\hline
SKDBERT$_6$ & None & 89.44 & 71.84 & 88.52 & 91.63 & 86.44 & 90.70 & 82.55 \\
\hline
\multirow{14}{*}{SKDBERT$_6$}
& T$_{01}$   & 84.61 (\textcolor{green}{$\downarrow$}) & 67.87 (\textcolor{green}{$\downarrow$})  & 88.78 (\textcolor{red}{$\uparrow$}) & 88.76 (\textcolor{green}{$\downarrow$}) & 84.63 (\textcolor{green}{$\downarrow$}) & 86.33 (\textcolor{green}{$\downarrow$}) & 74.87 (\textcolor{green}{$\downarrow$}) \\
& T$_{02}$   & 87.94 (\textcolor{green}{$\downarrow$}) & 67.15 (\textcolor{green}{$\downarrow$})  & 88.73 (\textcolor{red}{$\uparrow$}) & 89.68 (\textcolor{green}{$\downarrow$}) & 85.07 (\textcolor{green}{$\downarrow$}) & 86.20 (\textcolor{green}{$\downarrow$}) & 75.31 (\textcolor{green}{$\downarrow$}) \\
& T$_{03}$   & 87.08 (\textcolor{green}{$\downarrow$}) & 70.76 (\textcolor{green}{$\downarrow$})  & 88.95 (\textcolor{red}{$\uparrow$}) & 90.48 (\textcolor{green}{$\downarrow$}) & 86.08 (\textcolor{green}{$\downarrow$}) & 86.53 (\textcolor{green}{$\downarrow$}) & 76.21 (\textcolor{green}{$\downarrow$}) \\
& T$_{04}$   & 89.72 (\textcolor{red}{$\uparrow$}) & 68.95 (\textcolor{green}{$\downarrow$})  & \textbf{89.16} (\textcolor{red}{$\uparrow$}) & 92.43 (\textcolor{red}{$\uparrow$}) & 87.09 (\textcolor{red}{$\uparrow$}) & 89.57 (\textcolor{green}{$\downarrow$}) & 79.67 (\textcolor{green}{$\downarrow$}) \\
& T$_{05}$   & 89.73 (\textcolor{red}{$\uparrow$}) & 71.12 (\textcolor{green}{$\downarrow$})  & 88.81 (\textcolor{red}{$\uparrow$}) & 91.40 (\textcolor{green}{$\downarrow$}) & 87.77 (\textcolor{red}{$\uparrow$}) & 89.82 (\textcolor{green}{$\downarrow$}) & 79.84 (\textcolor{green}{$\downarrow$}) \\
& T$_{06}$   & 88.36 (\textcolor{green}{$\downarrow$}) & 69.31 (\textcolor{green}{$\downarrow$})  & 88.91 (\textcolor{red}{$\uparrow$}) & 92.43 (\textcolor{red}{$\uparrow$}) & 87.60 (\textcolor{red}{$\uparrow$}) & 90.24 (\textcolor{green}{$\downarrow$}) & 80.42 (\textcolor{green}{$\downarrow$}) \\
& T$_{07}$   & 89.46 (\textcolor{red}{$\uparrow$}) & \textbf{73.65} (\textcolor{red}{$\uparrow$}) & 88.81 (\textcolor{red}{$\uparrow$}) & 92.78 (\textcolor{red}{$\uparrow$}) & 88.64 (\textcolor{red}{$\uparrow$}) & 90.99 (\textcolor{red}{$\uparrow$}) & 81.93 (\textcolor{green}{$\downarrow$}) \\
& T$_{08}$   & 89.56 (\textcolor{red}{$\uparrow$}) & 71.48 (\textcolor{green}{$\downarrow$})  & 89.07 (\textcolor{red}{$\uparrow$}) & 92.32 (\textcolor{red}{$\uparrow$}) & 88.76 (\textcolor{red}{$\uparrow$}) & 90.90 (\textcolor{red}{$\uparrow$}) & 82.27 (\textcolor{green}{$\downarrow$}) \\
& T$_{09}$   & 89.85 (\textcolor{red}{$\uparrow$}) & 72.20 (\textcolor{red}{$\uparrow$}) & 89.02 (\textcolor{red}{$\uparrow$}) & 92.20 (\textcolor{red}{$\uparrow$})& 88.92 (\textcolor{red}{$\uparrow$}) & \textbf{91.51} (\textcolor{red}{$\uparrow$}) & 82.70 (\textcolor{red}{$\uparrow$}) \\
& T$_{10}$  & 89.59 (\textcolor{red}{$\uparrow$}) & 73.29 (\textcolor{red}{$\uparrow$}) & 89.06 (\textcolor{red}{$\uparrow$}) & 91.97 (\textcolor{red}{$\uparrow$}) & 88.88 (\textcolor{red}{$\uparrow$}) & 91.07 (\textcolor{red}{$\uparrow$}) & 82.86 (\textcolor{red}{$\uparrow$}) \\
& T$_{11}$  & 89.73 (\textcolor{red}{$\uparrow$}) & 71.84 (\textcolor{red}{$\uparrow$}) & 88.95 (\textcolor{red}{$\uparrow$}) & 92.32 (\textcolor{red}{$\uparrow$}) & 88.98 (\textcolor{red}{$\uparrow$}) & 91.27 (\textcolor{red}{$\uparrow$}) & 83.17 (\textcolor{red}{$\uparrow$}) \\
& T$_{12}$  & 89.13 (\textcolor{green}{$\downarrow$}) & 71.48 (\textcolor{green}{$\downarrow$})  & 89.02 (\textcolor{red}{$\uparrow$}) & \textbf{93.12} (\textcolor{red}{$\uparrow$}) & 88.90 (\textcolor{red}{$\uparrow$}) & 91.40 (\textcolor{red}{$\uparrow$}) & 82.83 (\textcolor{red}{$\uparrow$}) \\
& T$_{13}$  & \textbf{90.01} (\textcolor{red}{$\uparrow$}) & 72.92 (\textcolor{red}{$\uparrow$}) & 88.86 (\textcolor{red}{$\uparrow$}) & 92.09 (\textcolor{red}{$\uparrow$}) & 88.89 (\textcolor{red}{$\uparrow$}) & 91.16 (\textcolor{red}{$\uparrow$}) & 83.42 (\textcolor{red}{$\uparrow$}) \\
& T$_{14}$  & 89.50 (\textcolor{red}{$\uparrow$}) & 72.56 (\textcolor{red}{$\uparrow$}) & 88.86 (\textcolor{red}{$\uparrow$}) & 92.43 (\textcolor{red}{$\uparrow$}) & \textbf{89.03} (\textcolor{red}{$\uparrow$}) & 91.32 (\textcolor{red}{$\uparrow$}) & \textbf{83.46} (\textcolor{red}{$\uparrow$}) \\
\hline
\end{tabular}
\end{threeparttable}
\end{table}

\section{The Used Sampling Distributions for SKDBERT with Various Teacher Teams on The GLUE Benchmark}

\begin{figure}[!h]
	\begin{center}
		\includegraphics[width=0.44\textwidth]{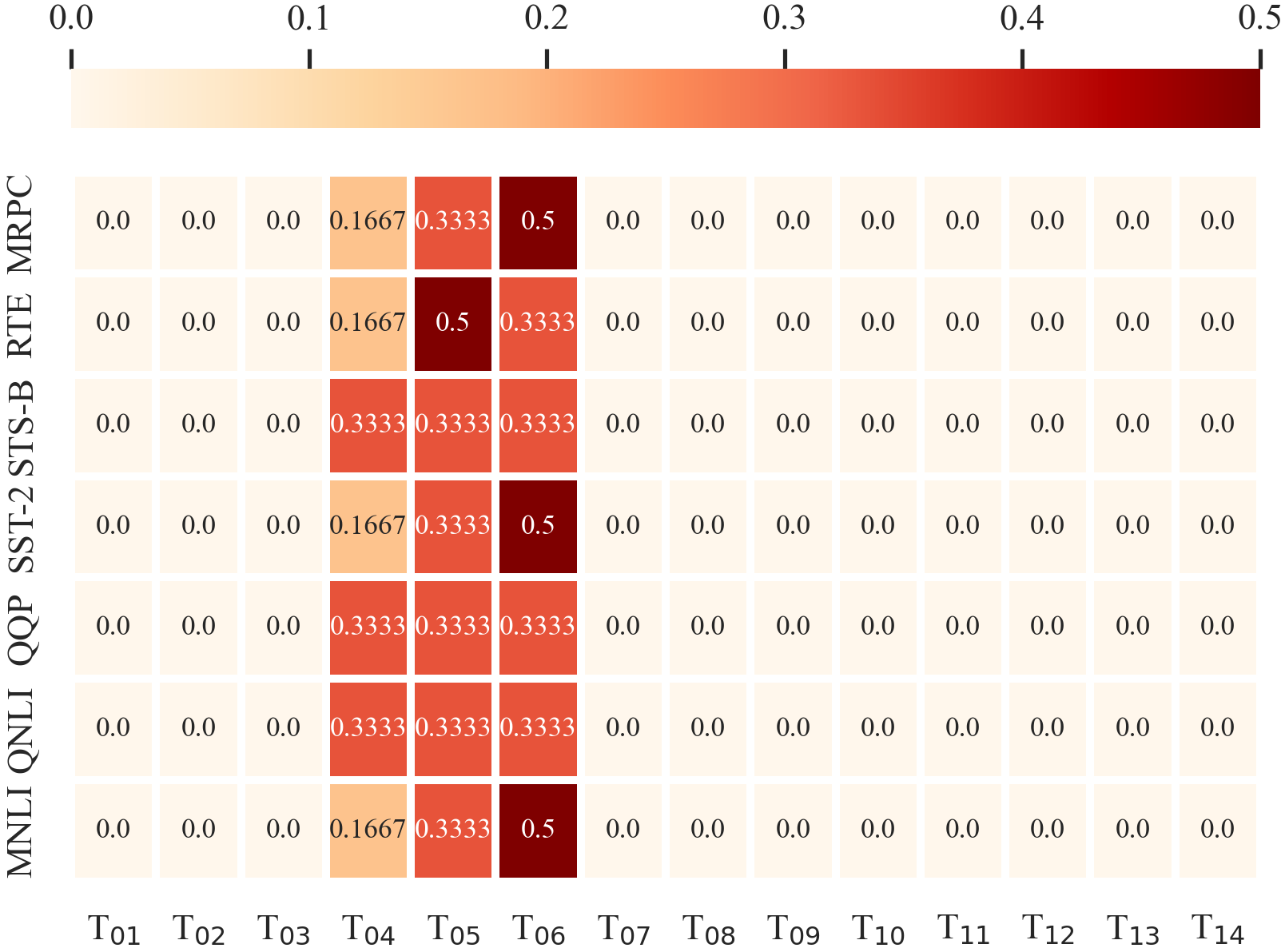}
	\end{center}
	\caption{The sampling distribution of SKDBERT$_4$ with the teacher team of T$_{04}$ to T$_{06}$ on the GLUE benchmark.}
  \label{fig:dis_4_1}
\end{figure}

\begin{figure}[!h]
	\begin{center}
		\includegraphics[width=0.44\textwidth]{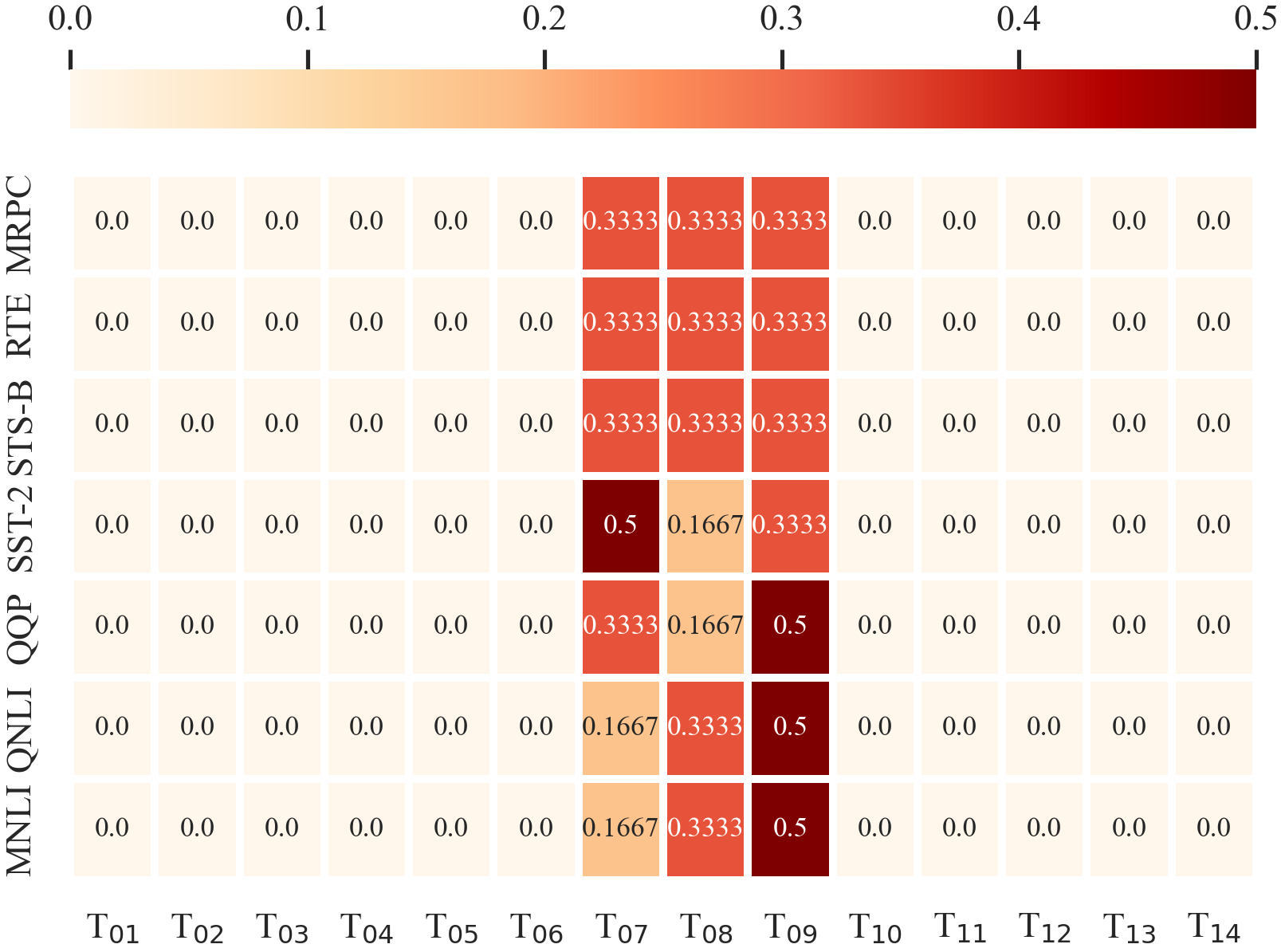}
	\end{center}
	\caption{The sampling distribution of SKDBERT$_4$ with the teacher team of T$_{07}$ to T$_{09}$ on the GLUE benchmark.}
  \label{fig:dis_4_2}
\end{figure}

\begin{figure}[!h]
	\begin{center}
		\includegraphics[width=0.44\textwidth]{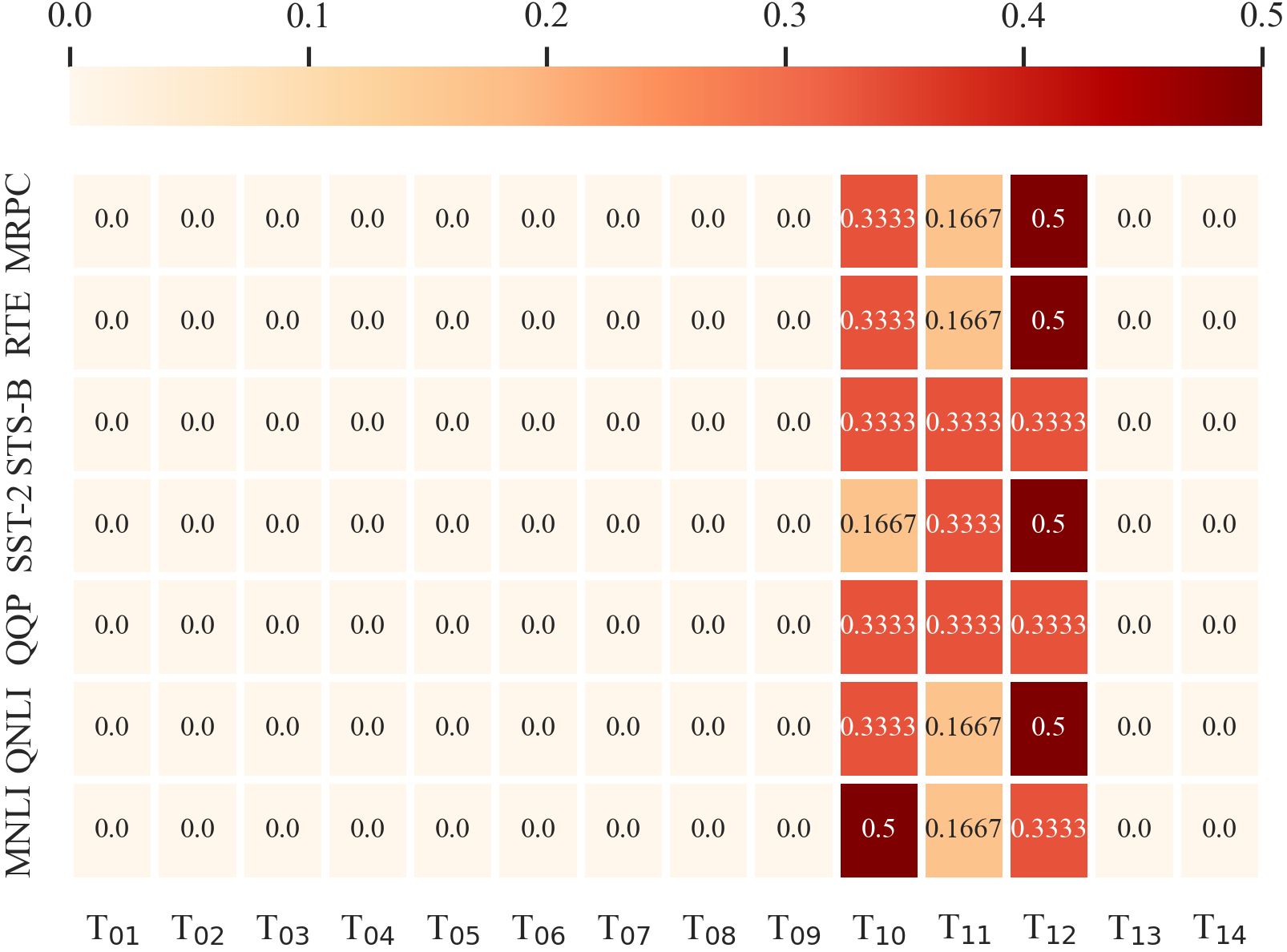}
	\end{center}
	\caption{The sampling distribution of SKDBERT$_4$ with the teacher team of T$_{10}$ to T$_{12}$ on the GLUE benchmark.}
  \label{fig:dis_4_3}
\end{figure}

\begin{figure}[!h]
	\begin{center}
		\includegraphics[width=0.44\textwidth]{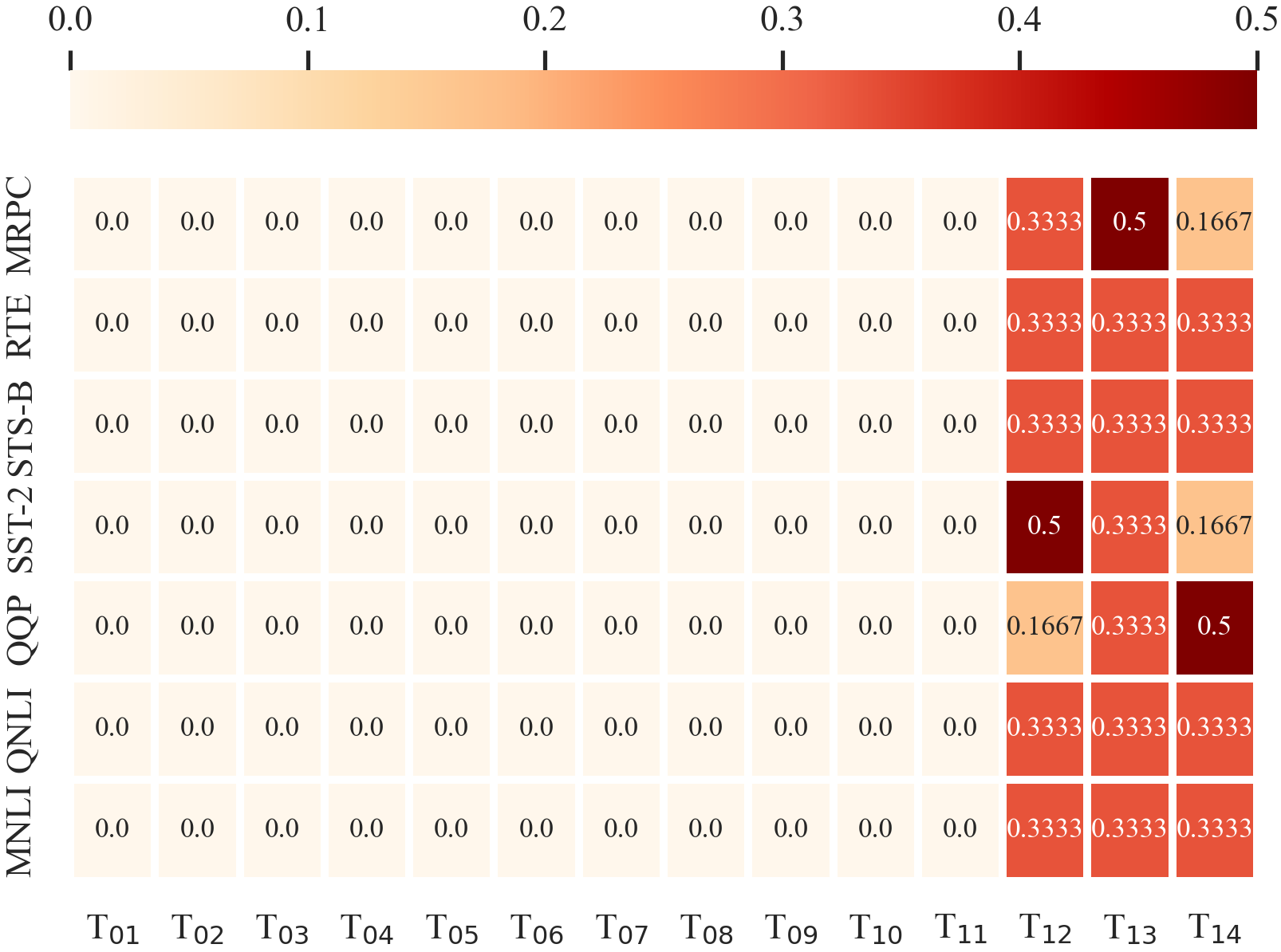}
	\end{center}
	\caption{The sampling distribution of SKDBERT$_4$ with the teacher team of T$_{12}$ to T$_{14}$ on the GLUE benchmark.}
  \label{fig:dis_4_4}
\end{figure}

\begin{figure}[!h]
	\begin{center}
		\includegraphics[width=0.44\textwidth]{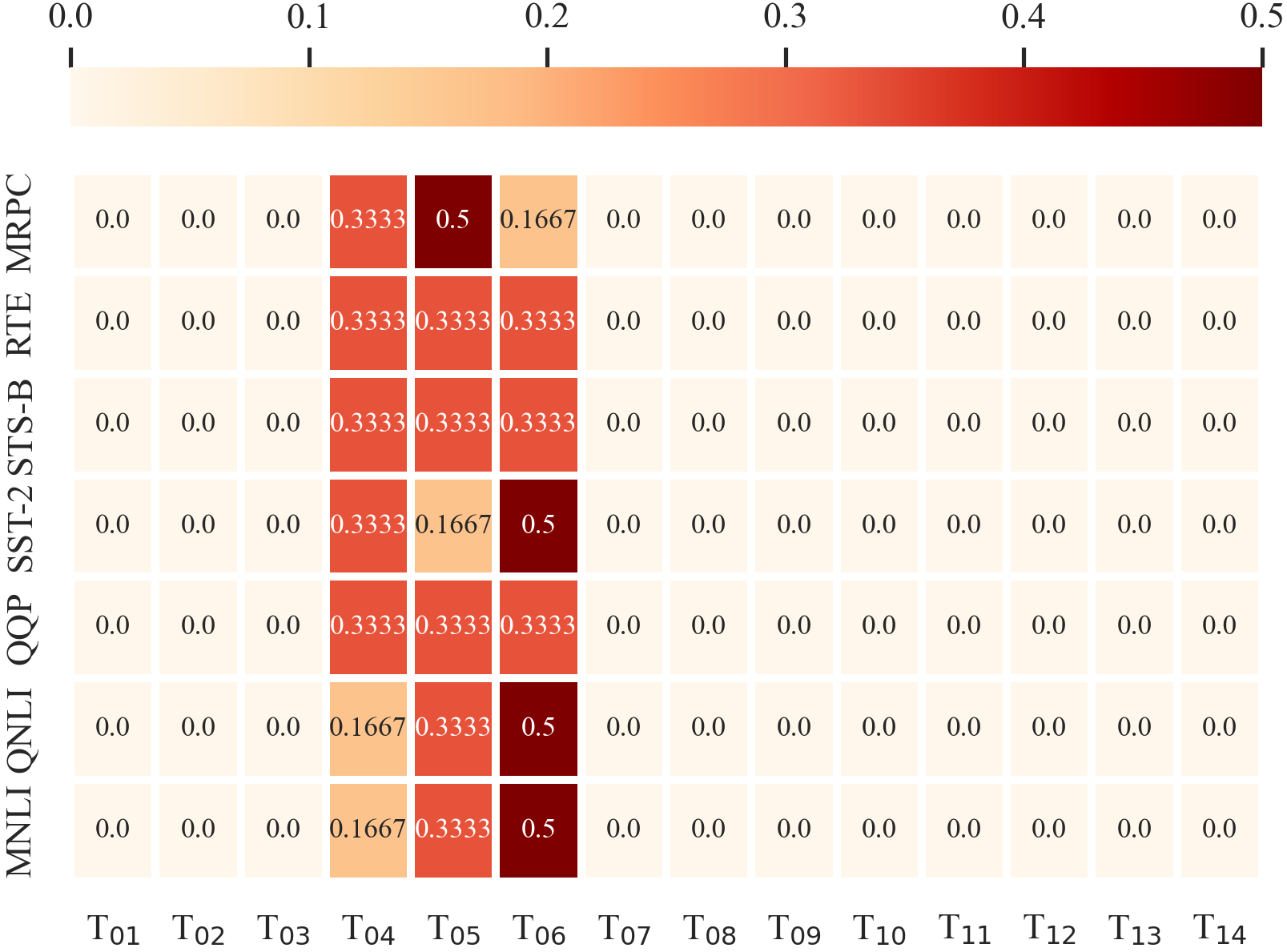}
	\end{center}
	\caption{The sampling distribution of SKDBERT$_6$ with the teacher team of T$_{04}$ to T$_{06}$ on the GLUE benchmark.}
  \label{fig:dis_6_1}
\end{figure}

\begin{figure}[!h]
	\begin{center}
		\includegraphics[width=0.44\textwidth]{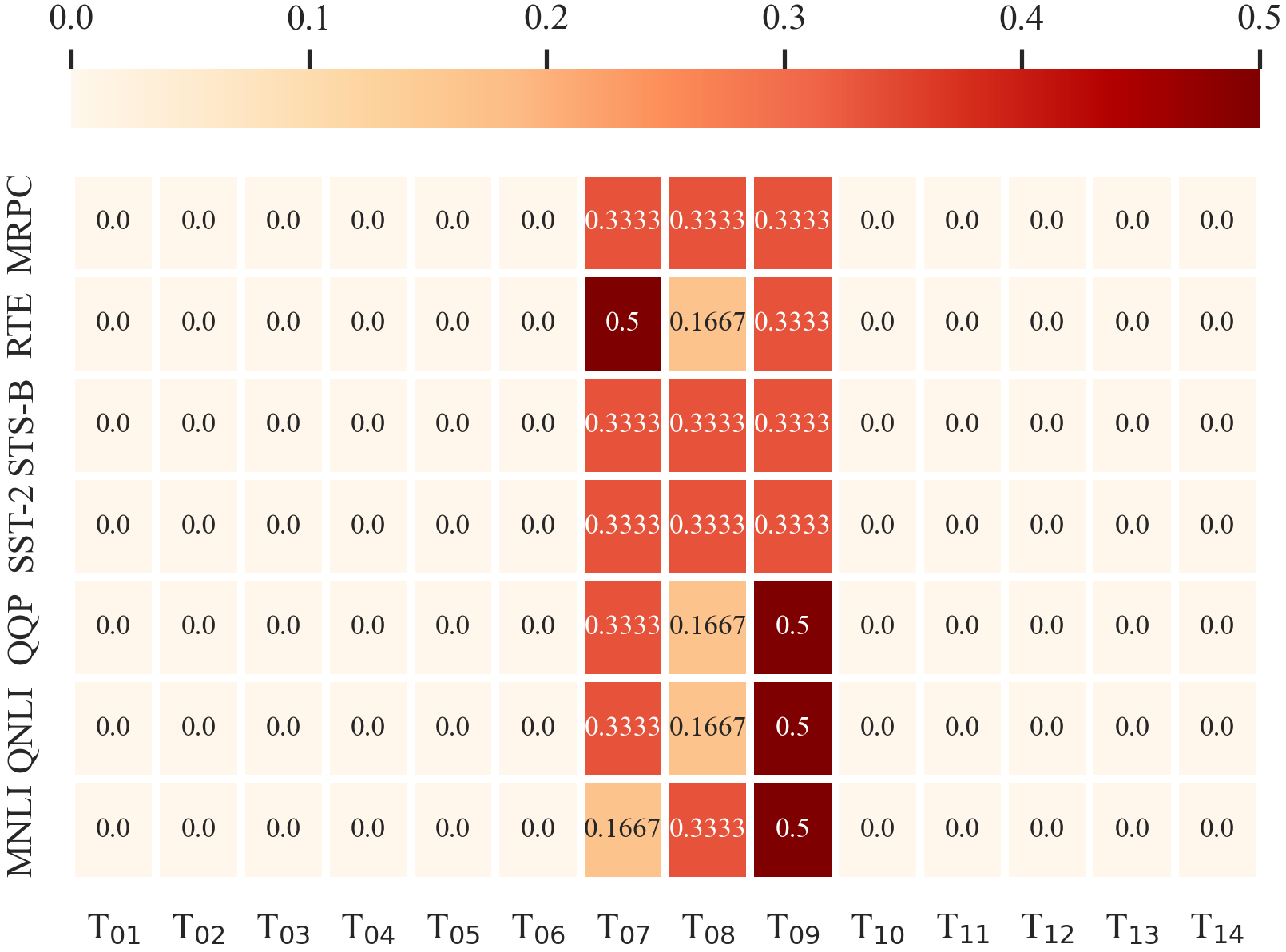}
	\end{center}
	\caption{The sampling distribution of SKDBERT$_6$ with the teacher team of T$_{07}$ to T$_{09}$ on the GLUE benchmark.}
  \label{fig:dis_6_2}
\end{figure}

\begin{figure}[!h]
	\begin{center}
		\includegraphics[width=0.44\textwidth]{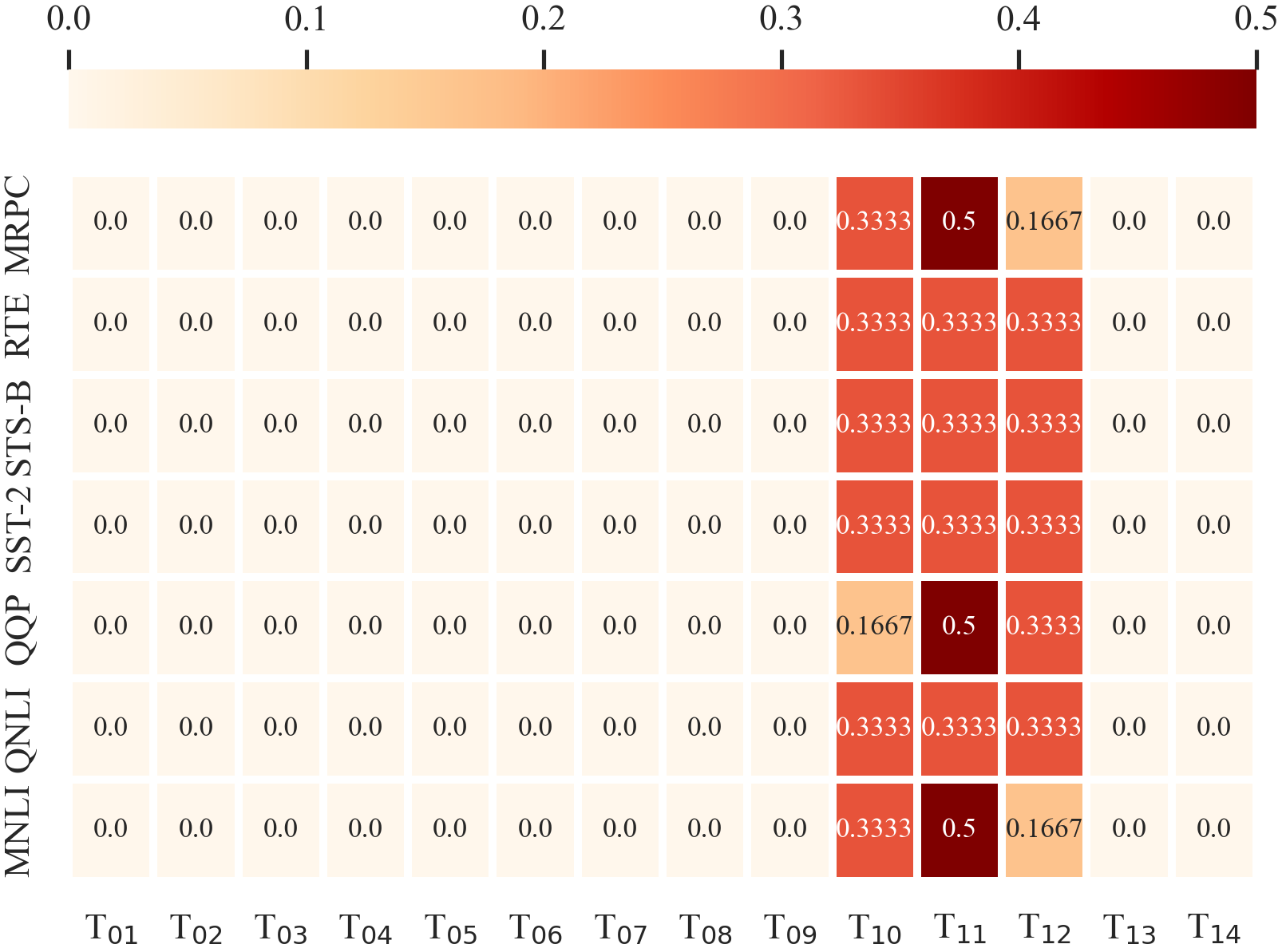}
	\end{center}
	\caption{The sampling distribution of SKDBERT$_6$ with the teacher team of T$_{10}$ to T$_{12}$ on the GLUE benchmark.}
  \label{fig:dis_6_3}
\end{figure}

\begin{figure}[!h]
	\begin{center}
		\includegraphics[width=0.43\textwidth]{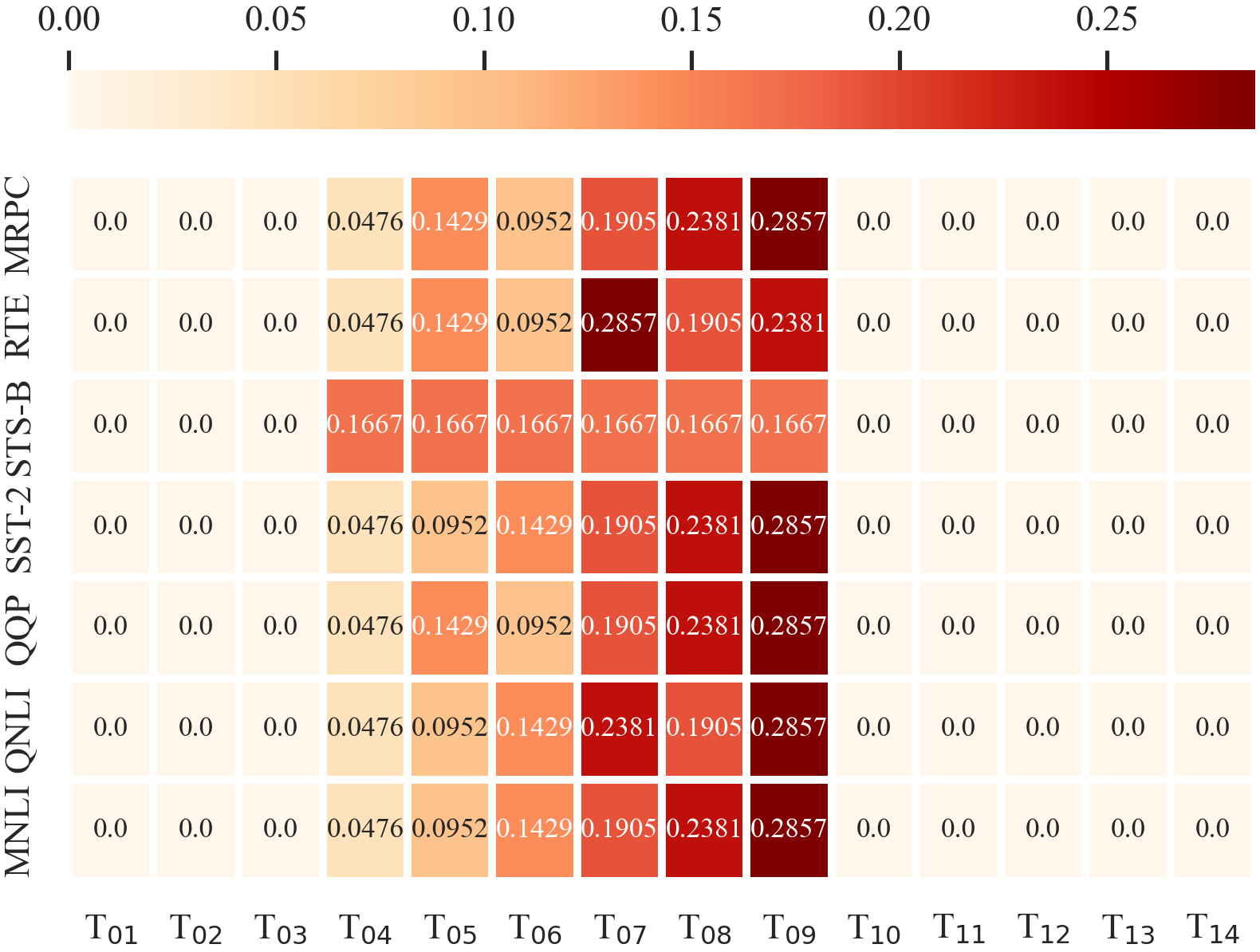}
	\end{center}
	\caption{The sampling distribution of SKDBERT$_6$ with the teacher team of T$_{04}$ to T$_{09}$ on the GLUE benchmark.}
  \label{fig:dis_6_4}
\end{figure}

\begin{figure}[!h]
	\begin{center}
		\includegraphics[width=0.44\textwidth]{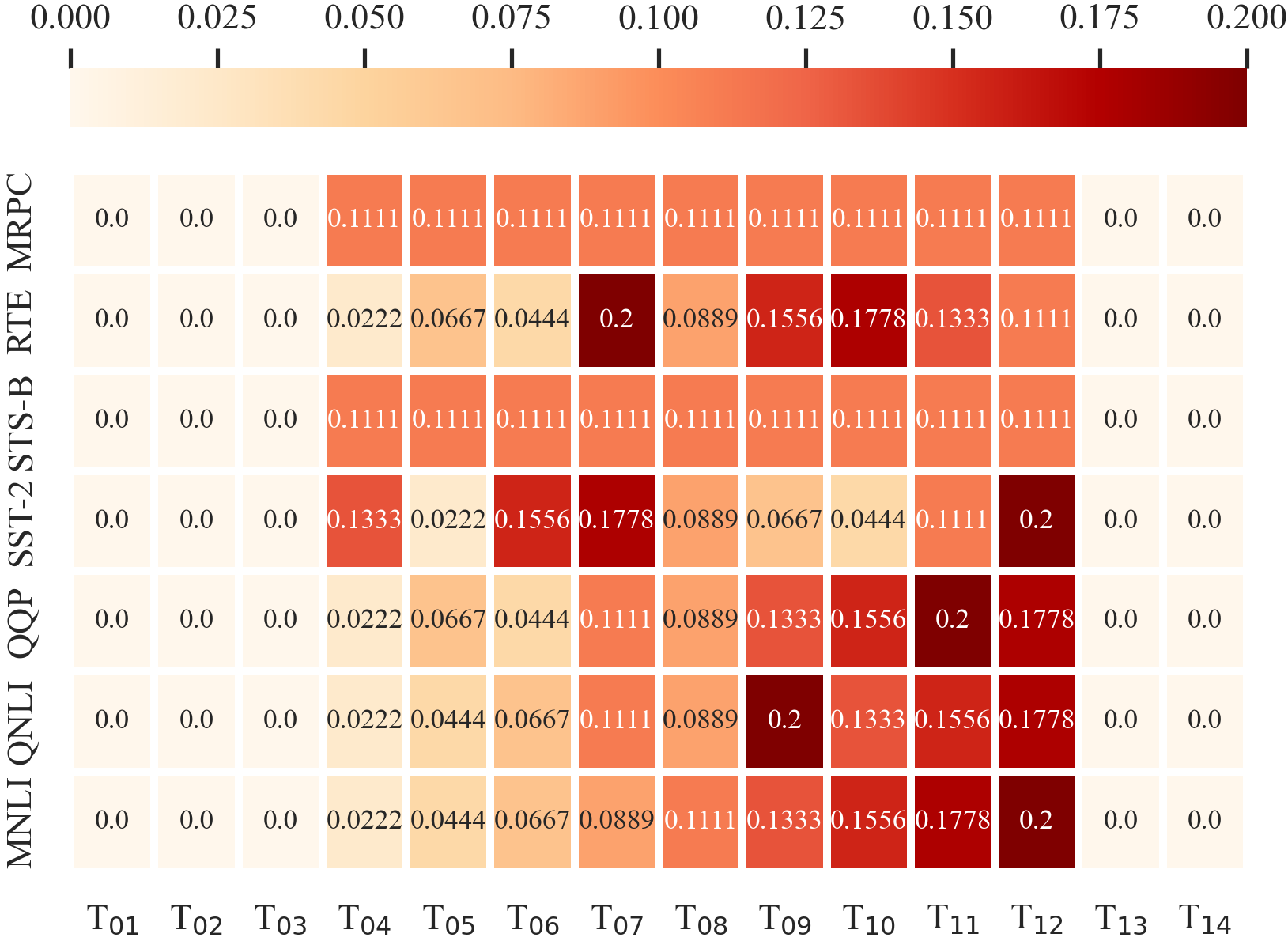}
	\end{center}
	\caption{The sampling distribution of SKDBERT$_6$ with the teacher team of T$_{04}$ to T$_{12}$ on the GLUE benchmark.}
  \label{fig:dis_6_5}
\end{figure}

\begin{figure}[!h]
	\begin{center}
		\includegraphics[width=0.43\textwidth]{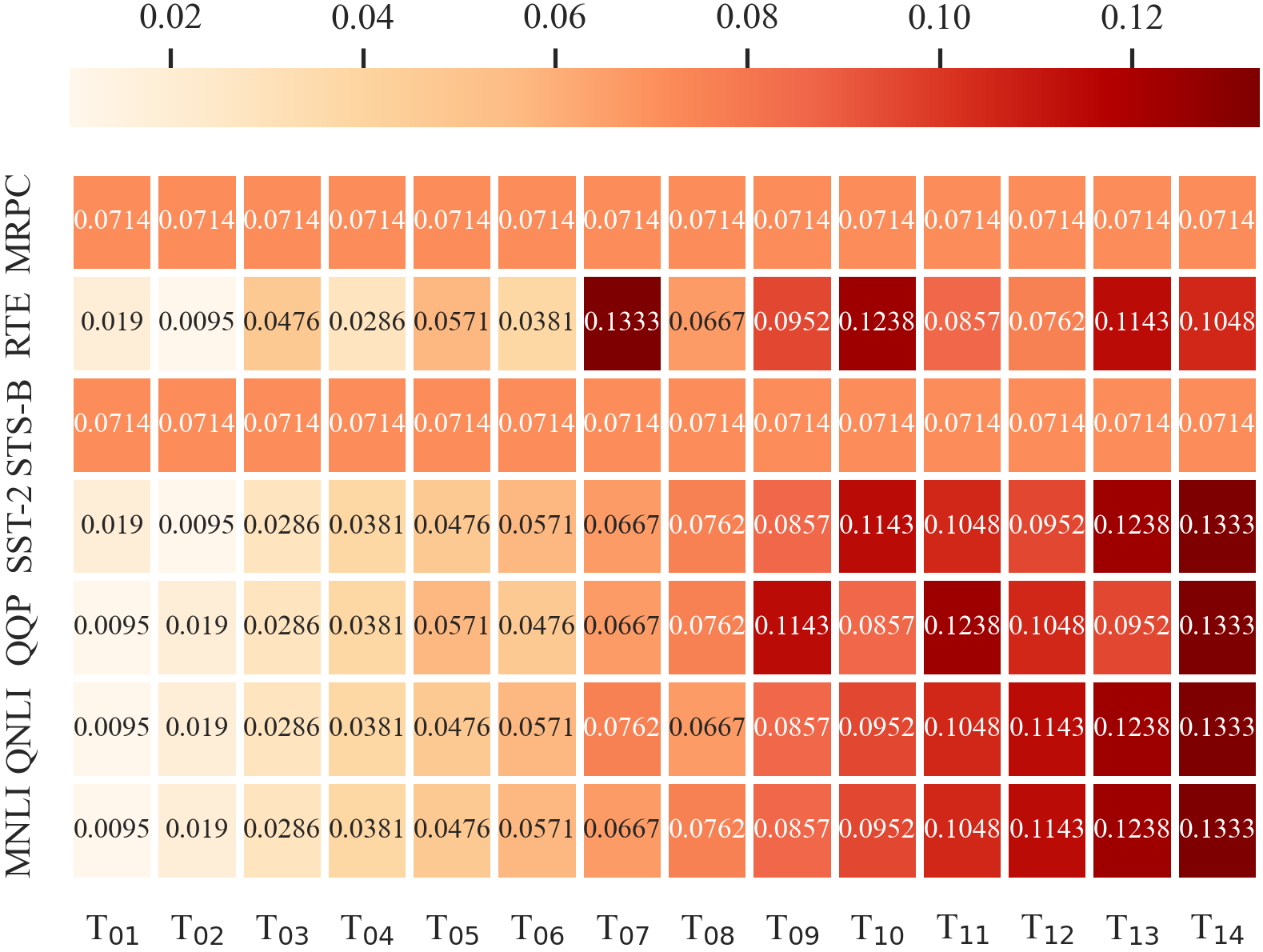}
	\end{center}
	\caption{The sampling distribution of SKDBERT$_6$ with the teacher team of T$_{01}$ to T$_{14}$ on the GLUE benchmark.}
  \label{fig:dis_6_6}
\end{figure}

\begin{figure}[!h]
	\begin{center}
		\includegraphics[width=0.43\textwidth]{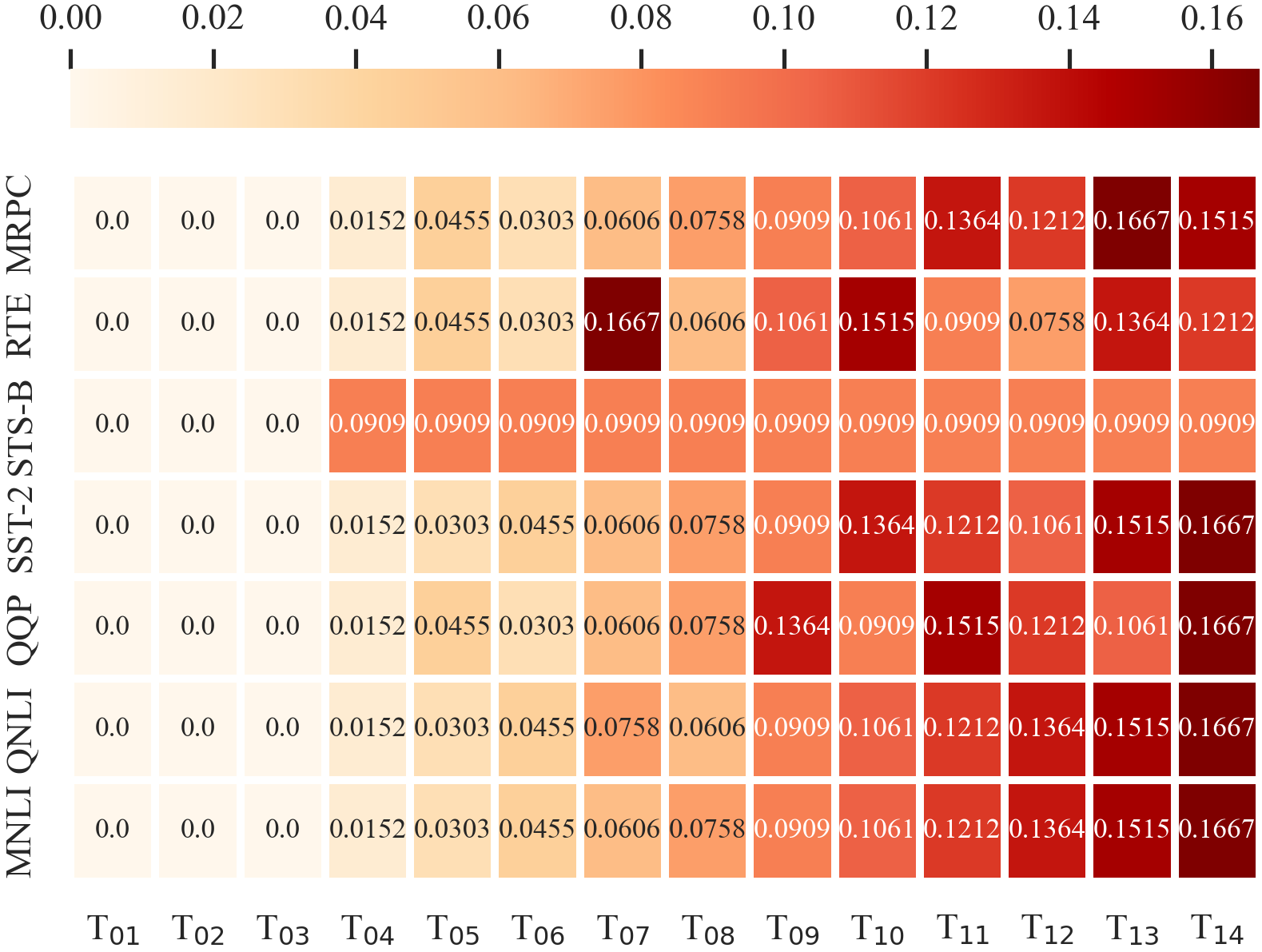}
	\end{center}
	\caption{The sampling distribution of SKDBERT$_6$ with the teacher team of T$_{04}$ to T$_{14}$ on the GLUE benchmark.}
  \label{fig:dis_6_7}
\end{figure}

\begin{figure}[!h]
	\begin{center}
		\includegraphics[width=0.43\textwidth]{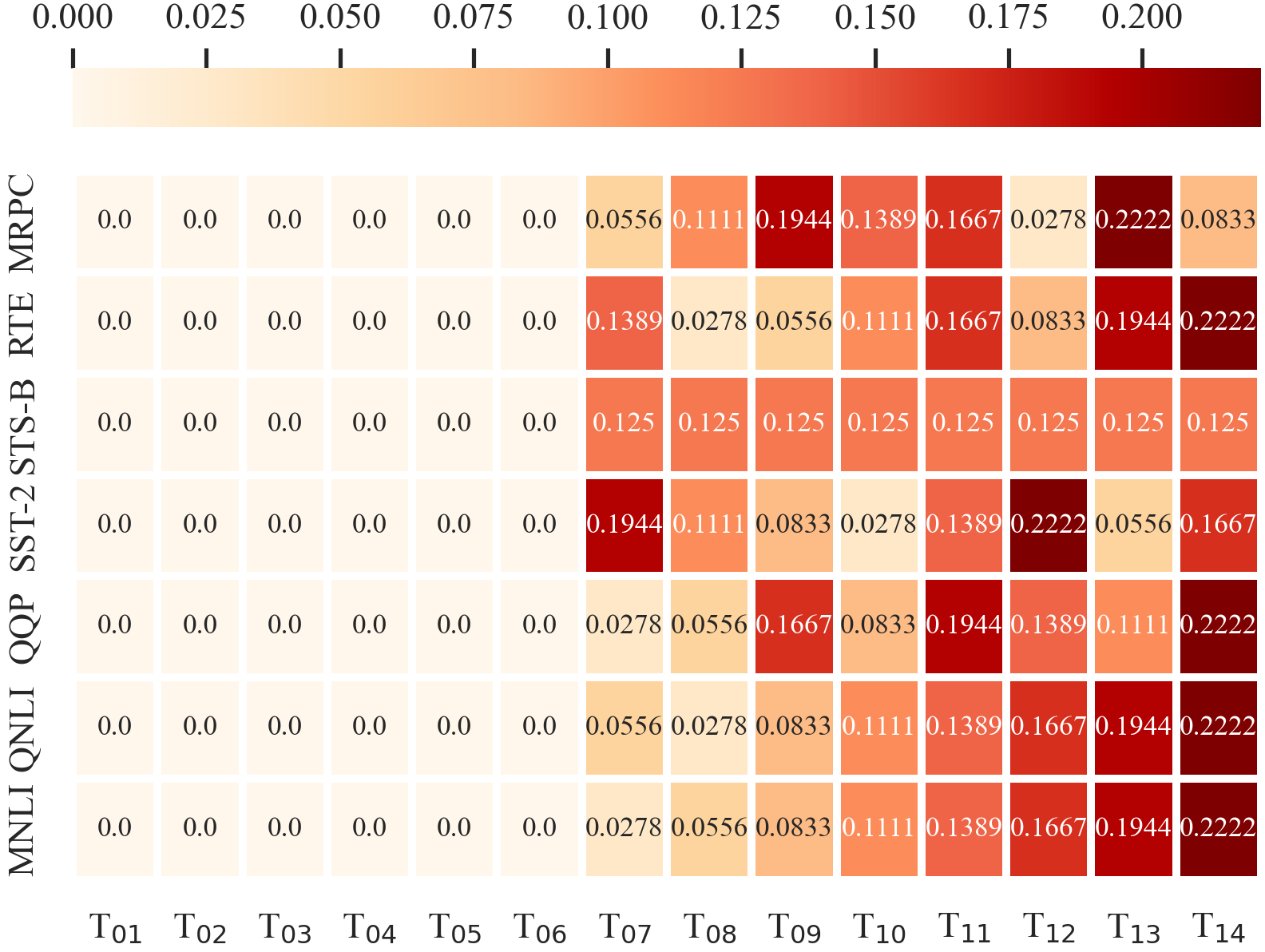}
	\end{center}
	\caption{The sampling distribution of SKDBERT$_6$ with the teacher team of T$_{07}$ to T$_{14}$ on the GLUE benchmark.}
  \label{fig:dis_6_8}
\end{figure}

\begin{figure}[!h]
	\begin{center}
		\includegraphics[width=0.43\textwidth]{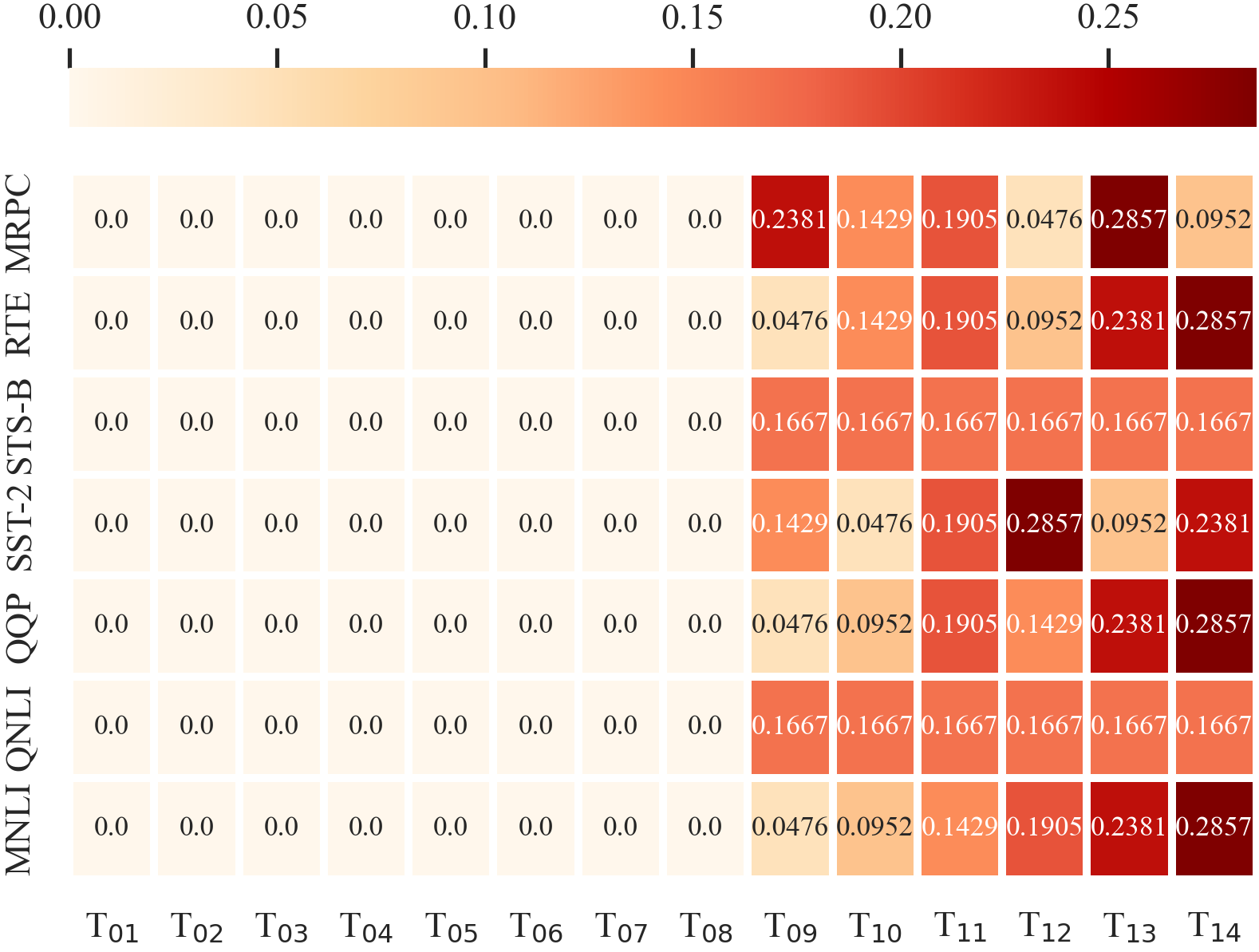}
	\end{center}
	\caption{The sampling distribution of SKDBERT$_6$ with the teacher team of T$_{09}$ to T$_{14}$ on the GLUE benchmark.}
  \label{fig:dis_6_9}
\end{figure}

\begin{figure}[!h]
	\begin{center}
		\includegraphics[width=0.44\textwidth]{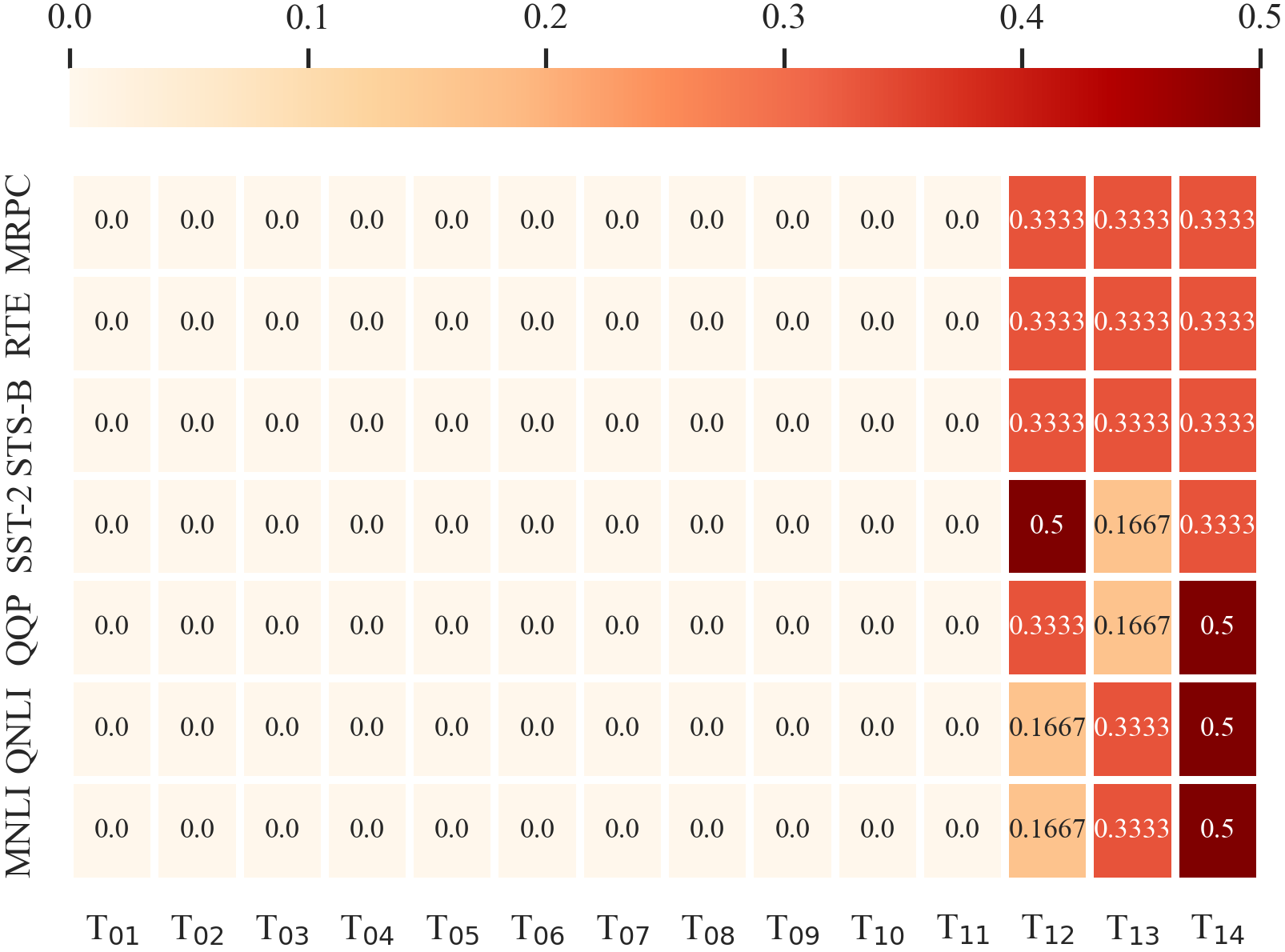}
	\end{center}
	\caption{The sampling distribution of SKDBERT$_6$ with the teacher team of T$_{12}$ to T$_{14}$ on the GLUE benchmark.}
  \label{fig:dis_6_11}
\end{figure}

\begin{figure}[!h]
	\begin{center}
		\includegraphics[width=0.43\textwidth]{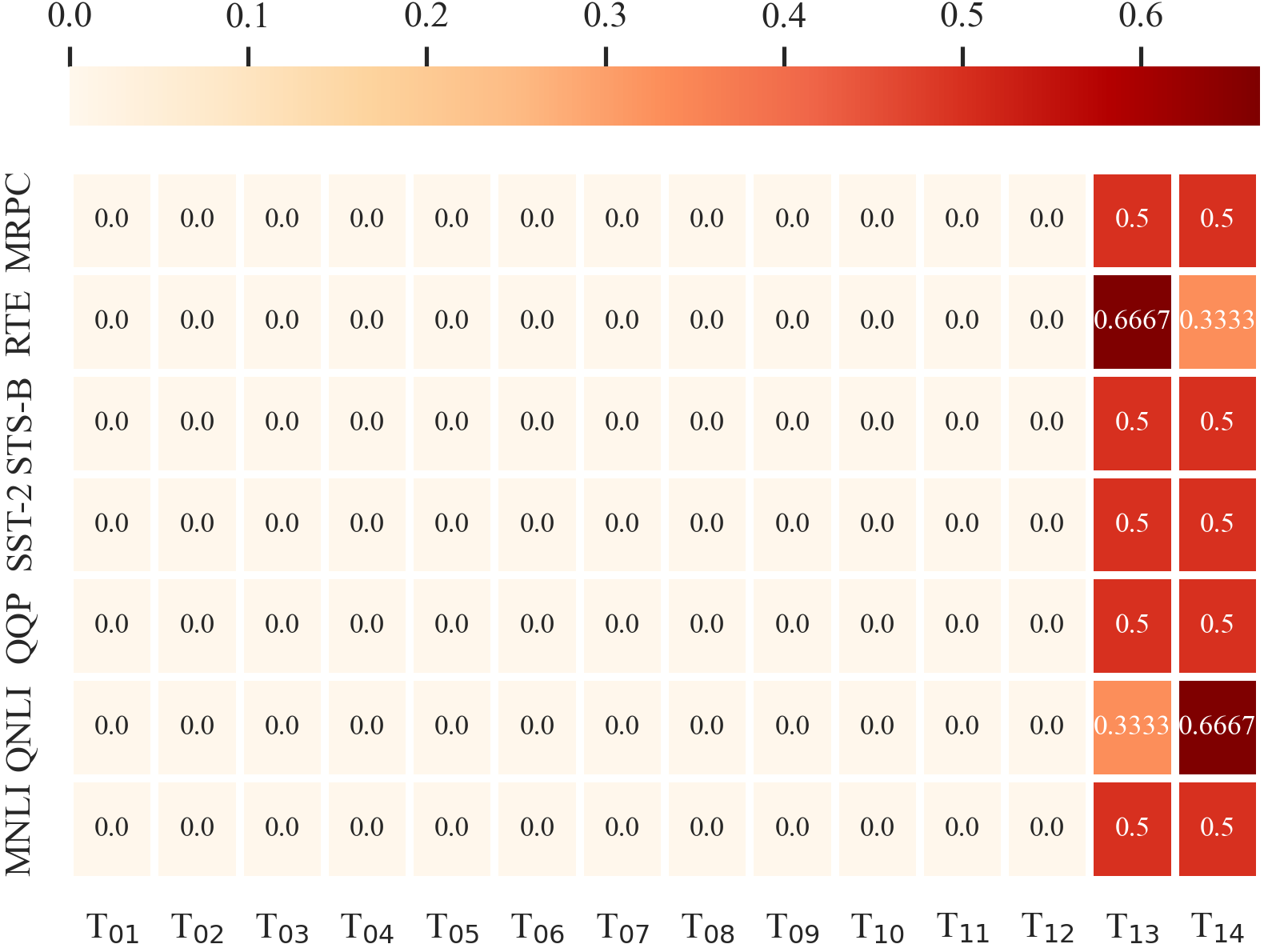}
	\end{center}
	\caption{The sampling distribution of SKDBERT$_6$ with the teacher team of T$_{13}$ to T$_{14}$ on the GLUE benchmark.}
  \label{fig:dis_6_12}
\end{figure}

We implement extensive ablation experiments for SKDBERT using various teacher teams: \{T$_{04}$-T$_{06}$, T$_{07}$-T$_{09}$, T$_{10}$-T$_{12}$,  T$_{12}$-T$_{14}$\} for SKDBERT$_4$ and \{T$_{04}$-T$_{06}$, T$_{07}$-T$_{09}$, T$_{10}$-T$_{12}$, T$_{04}$-T$_{09}$, T$_{04}$-T$_{12}$, T$_{01}$-T$_{14}$, T$_{04}$-T$_{14}$, T$_{07}$-T$_{14}$, T$_{09}$-T$_{14}$, T$_{12}$-T$_{14}$, T$_{13}$-T$_{14}$\} for SKDBERT$_6$. We show the used distributions for the above teacher teams on various tasks in Figure \ref{fig:dis_4_1} to \ref{fig:dis_6_12}. Particular, 0 indicates that the teacher models are not employed for stochastic knowledge distillation. 

\bibliography{references.bib}

\end{document}